%% file: arxiv.tex
\documentclass[10pt,twocolumn,letterpaper]{article}

\usepackage{iccv}
\usepackage{times}
\usepackage{epsfig}
\usepackage{graphicx}
\usepackage{amsmath}
\usepackage{amssymb}
\usepackage{mathtools}

\usepackage{booktabs}
\usepackage{algorithm}
\usepackage{lipsum}
\usepackage{url}

\usepackage{tabularx}
\usepackage{multirow}
\usepackage{xcolor,colortbl}
\usepackage{wrapfig}

\newcommand{\PAR}[1]{\vskip4pt \noindent {\bf #1~}}

\usepackage[pagebackref=true,breaklinks=true,colorlinks,bookmarks=false]{hyperref}

\iccvfinalcopy 

\def\iccvPaperID{12389} 

\def \lid {LiT decoder\xspace}

\begin{document}

\title{A Study of Autoregressive Decoders for Multi-Tasking in Computer Vision}

\newcommand{\authsep}{\;\;\;\;}

\author{
Lucas Beyer$^{\star\dagger}_1$ \authsep
Bo Wan$^{\star}_{1,3}$ \authsep
Gagan Madan$^{\star}_2$ \authsep
Filip Pavetic$^{\star}_1$ \authsep
Andreas Steiner$^{\star}_1$ \\
Alexander Kolesnikov$_1$ \authsep
Andr\'e Susano Pinto$_1$ \authsep
Emanuele Bugliarello$_4$ \\
Xiao Wang$_1$ \authsep
Qihang Yu$_{2,5}$ \authsep
Liang-Chieh Chen$_2$ \authsep
Xiaohua Zhai$^{\star\dagger}_1$ \\
\\
Google Research}

\maketitle

{\let\thefootnote\relax\footnote{
{\hspace{-2em}$^{\star}$ Equal technical contribution. $^\dagger$ LB and XZ started and led the project. \\
$^1$ Google Research, Brain Team.
$^2$ Google Research.\\
$^{3,4,5}$ work done during Google internship, while doing PhD at:\\
$^{\,\,3}$KU Leuven, $^4$University of Copenhagen, $^5$Johns Hopkins University.
}}}

\maketitle

\vspace{-1em}%
\begin{abstract}
There has been a recent explosion of computer vision models which perform many tasks and are composed of an image encoder (usually a ViT) and an autoregressive decoder (usually a Transformer).
However, most of this work simply presents one system and its results, leaving many questions regarding design decisions and trade-offs of such systems unanswered. 
In this work, we aim to provide such answers.
We take a close look at autoregressive decoders for multi-task learning in multimodal computer vision, including classification, captioning, visual question answering, and optical character recognition.
Through extensive systematic experiments, we study the effects of task and data mixture, training and regularization hyperparameters, conditioning type and specificity, modality combination, and more.
Importantly, we compare these to well-tuned single-task baselines to highlight the cost incurred by multi-tasking.
A key finding is that a small decoder learned on top of a frozen pretrained encoder works surprisingly well.
We call this setup locked-image tuning with decoder (\lid).
It can be seen as teaching a decoder to interact with a pretrained vision model via natural language.
\end{abstract}

\section{Introduction}\label{sec:intro}

In recent years, deep learning models have started to attain reasonably satisfying performance on many individual tasks across computer vision~\cite{Krizhevsky12imagenet,Dosovitskiy21AnII,carion20detr,li22explore,strudel2021segmenter,chen22asss,yang22temporally}.
At the same time, these models are becoming increasingly similar across tasks, mostly rallying behind the Transformer architecture~\cite{Dosovitskiy21AnII,liu21swin,wang2021not,wang21pvt,Heo2021rethinking,touvron21training}.
This convergence has led to a significantly increased interest in training a single model that can handle all kinds of computer vision tasks with as little task-specific architectural design as possible~\cite{yuan21florence,wang22ofa,yu2022coca,chen2022unified,zou2022xdecoder,lu23unifiedio}.
Specifically, an encoder-decoder Transformer with an autoregressive decoder is quickly establishing itself as the model of choice, since it allows reusing pretrained image encoders and the decoder can efficiently model joint distributions over output sequences.
The recent emergence of work turning classic dense-output tasks such as detection and segmentation into sequence modeling tasks further cements this choice~\cite{kolesnikov22uvim,zou2022xdecoder,lu23unifiedio}.
However, the vast majority of current work simply presents a complete, usually large, system and its capabilities without providing insights on the benefits and pitfalls of multi-tasking with autoregressive decoders.
This paper aims to fill that gap.

Specifically, we take a strong pretrained vision encoder which is able to extract high-level knowledge from an image, and then train a multi-task autoregressive decoder on top of it while keeping the encoder frozen.
We name this setup locked-image tuning with decoder (\lid).
In this setup, we first answer the crucial question of how one multi-task \lid performs compared to heavily tuned single-task \lid baselines on each task, in a controlled setup.
We demonstrate the effectiveness of task-conditioning the decoder (\ie, telling it which task it is supposed to solve) and the failure modes of not doing so.
We then study the relationship between the number of tasks and the decoder's capacity, uncovering two different behavioral modes.
We also investigate the transfer of ``skills'' across tasks in a controlled way.
Finally, we dig into the effect of different datasets mixture strategies, and many more questions, all in an explicitly controlled setup and with varying types of tasks in order to draw general take-aways. 

Perhaps, some of the more surprising findings are (i) that multi-task \lid seems to require \emph{less} hyperparameter tuning than single-task \lid, because (ii) even the addition of unrelated tasks is beneficial, and (iii) that a two-layer decoder on top of a frozen backbone works well.

\section{Related work}\label{sec:relwork}

The literature on multi-task learning is vast and has been frequently surveyed~\cite{caruana1998multitask,zhang17survey,ruder17survey,Gong19Comparison,Crawshaw20multitask,Vandenhende22multi}.
Recent work~\cite{kurin22in} suggests that using a simple method, while regularizing the training and augmenting data properly, may be more important for good multi-task performance than using more advanced methods.
Hence, we focus our study on this setting: we employ a simple encoder-decoder setup and study the effects of various design decisions. 

Even when restricted to computer vision, the field is extensive.
Various sub-fields have emerged, concentrating on classification~\cite{Kuang17deepmulti,requeima2019cnaps,Alam21MEDIC,shen2022association}, dense outputs~\cite{Kokkinos17UberNet,Zamir2018Taskonomy,girdhar2022omnivore,Bhattacharjee22mult,ye22invpt}, or image--language tasks~\cite{li19visualbert,lu2019vilbert,su2020vlbert,wang21ufo,bugliarello21unmasked,li21albef,kim21vilt,li2022blip,bugliarello22iglue,chen22pali,wang22beitv3,wang22simvlm,piergiovanni2022pre}.
The latter are recently overtaking the field, with dense and structured output tasks being cast as language~\cite{lu20twelve,wang22ofa,chen22pix2seq,kolesnikov22uvim,lu23unifiedio,zou2022xdecoder}.
Hence, we focus our study on this ascending setting.

Different from most aforementioned work, we do not introduce any novel method.
Instead, we concentrate on understanding the effects of the most promising method for unifying all computer vision tasks: autoregressive decoders.
Another important distinction is that, since we focus on multi-task learning, we do neither consider zero-shot transfer~\cite{zhu22uni,chen22pali,wang22git,wang22beitv3,zou2022xdecoder}, nor finetuning on individual tasks~\cite{chen22pali,wang22git}.
However, we do highlight that the very simple setup of using a frozen, pretrained encoder with a very small decoder seems to be surprisingly effective.

\section{Background}\label{sec:background}

We provide a brief introduction to autoregressive decoders, the familiar reader may safely skip this section.

\makeatletter
\newcommand{\@giventhatstar}[2]{\ensuremath{\left(#1\,\middle|\,#2\right)}}
\newcommand{\@giventhatnostar}[3][]{\ensuremath{#1(#2\,#1|\,#3#1)}}
\newcommand{\giventhat}{\@ifstar\@giventhatstar\@giventhatnostar}
\makeatother

In general, a computer vision model can be defined as function that produces an output $y$ (such as class name, caption, answer, segmentation maps, \etc) for a given input image $x$ and optionally other auxiliary inputs.
One approach to this is to use a probabilistic generative model: a model that learns the function $p\giventhat{y}{x}$ and is able to compute the probability of (to score) any given $y$ for a given input. At inference time, such a model can be used to select (to decode) one concrete instance from all of its output space.

For single-variable outputs, such as classification tasks, this is simple.
We commonly compute the probability of each and every possible output class and then select the one achieving the highest value. 
However, for more complex outputs, $y$ is typically composed of multiple variables $y_1, y_2, \dots$, for example each $y_i$ could be a letter, a word, a segmented pixel or patch, a detection box, \etc.
This poses some problems. 
First, due to the cardinality of the output space, it is no longer viable to explicitly score every possible output.
Second, the outputs often have to be coherent and it is rarely sufficient to assume that output variables $y_i$ are independent from one another.
Thus, in practice, one requires a more implicit model of a \emph{joint} output distribution.

Autoregressive decoder models tackle this by using the product rule to \emph{exactly} rewrite $p\giventhat{y_1, y_2, \dots}{x} = p\giventhat{y_1}{x} p\giventhat{y_2}{y_1, x} \dots$ which highlights a way to tackle the task:
learn a model which takes as input the image $x$ and predicts a distribution over the first output value, another model which takes as input the image $x$ and a first output value $y_1$ and predicts a distribution over second output value $y_2$, and so forth.
Transformer models~\cite{Vaswani17attention} with a causal self-attention (where tokens can only attend to previous tokens) are efficient autoregressive decoders.
In particular, a single set of parameters and a single pass can be used to run all those predictors in parallel on an output sequence at training time.
At inference time, however, they still exhibit a sequential nature of only being able to select a $y_2$ after a concrete $y_1$ has been sampled.
Different strategies for constructing such sequences are discussed in Section~\ref{sec:results:decoding}.

Another interesting property is that autoregressive Transformer models provide significant flexibility in inputs.
In particular, two main methods exist:
cross-attention with an encoded input $x$, which we use to provide a representation of the input image in this work; and prefixing (prompting) the output sequence, which we use to both provide task conditioning and extra input information such as an input question for visual question answering tasks~\cite{antol15vqa,Goyal17vqav2,agrawal23reassessing}.
 
\section{Experimental setup}\label{sec:setup}

\input{tbls/results_core}

We explore multi-task learning with various types of computer vision tasks while using a single autoregressive decoder with a pretrained image encoder.
In order to isolate any studied effects to the autoregressive decoder, we use a frozen pretrained encoder in most experiments which we call \lid, but ablate the effect of this in Section~\ref{sec:results:nonfrozen}.
The encoder is a ViT-B/16~\cite{Dosovitskiy21AnII} that was pretrained on WebLI~\cite{chen22pali} using a sigmoid-based contrastive loss~\cite{siglip}. It achieves $69.9\%$ accuracy in zero-shot ImageNet-1k~\cite{Olga15ILSVRC} and $33.8\%$ averaged text-to-image retrieval results on the multilingual XM3600~\cite{xm3600}.

We include a set of $10$ representative tasks covering classification (CLS)~\cite{Olga15ILSVRC,zhou2017places,bossard14food,cheng17resisc,parkhi12a}, captioning (CAP)~\cite{chen15coco,young14flickr}, optical character recognition (OCR)~\cite{Mishra19ocrvqa} and question-answering (QA)~\cite{Goyal17vqav2,hudson19gqa}. We provide an overview of the datasets in the appendix.
We chose them to be easily accessible, cover specialized, fine-grained, and broad categories, and a wide range in dataset size.
We remove all images in the pretraining data that are near-duplicates of images from all the tasks we consider~\cite{kolesnikov2020bit}, in order to avoid any potentially misleading findings.

For training, we use ScalingViT-AdaFactor~\cite{zhai22scaling} which behaves like AdamW~\cite{Loshchilov19adamw} but is more memory efficient.
We use the multilingual SentencePiece tokenizer~\cite{kudo2018sentencepiece} from \cite{xue-etal-2021-mt5} with a vocabulary size 250{,}000 for all experiments, even English-only ones.
This slightly reduces performance across the board on typical English-only datasets.
The decoder is a standard Base Transformer model with an autoregressive mask on the self-attention, we only vary its depth in some experiments.
We restrict the sequence length to 64 tokens, covering the vast majority of lengths across tasks while being reasonably efficient compute-wise.
For single-task training, we use fewer tokens when possible; this is especially the case for classification.
Since we do not perform zero-shot transfer, we do not use ``A picture of'' prompts but stick to raw labels, with a per-task prefix (Sec.~\ref{sec:setup:multitask}).

At inference time, we default to greedy decoding for the classification and QA tasks and beam search with 4 beams for captioning and OCR.
This means that at inference time the decoder model is able to output any string even for classification tasks. 
We evaluate classification tasks, OCR and QA tasks using accuracy, only counting exact string matches as correct.
For captioning tasks we report CIDEr~\cite{vedantam2015cider} scores.

\subsection{Single-task baselines}\label{sec:setup:baselines}
As we aim to confidently evaluate the benefit or harm of multi-task learning over training a separate model for each task, we make sure to heavily tune our single-task baselines.
In fact, we spent considerably more time and compute tuning those than the multi-task setup.
We evaluate two kinds of single-task baselines.
A single-task \lid with the same ViT-B/16 encoder and an autoregressive Base Transformer decoder of varying depth and, for the classification tasks, linear, MLP, and self-attention~\cite{lee19set} classifiers on top of the same frozen backbone.
We sweep over the product of many hyperparameter values, including learning-rate, weight-decay, epochs, model size, dropout, label smoothing, crop augmentation, and randaugment~\cite{cubuk19randaug}.
We select the best performing model on a held-out validation set and re-train it with three different random seeds in order to report an average score on the test set.
Detailed settings are provided in the appendix.

\subsection{Multi-task \lid training}\label{sec:setup:multitask}
While we separately evaluate most design choices, the common starting point consists of creating minibatches randomly filled with examples from all tasks, sampled according to the task's sizes (which vary from over 1\,M images to just 3{,}000) such that one multi-task epoch roughly corresponds to one epoch on all of the tasks.
We condition the model on each task by prefixing a single-token task prompt followed by a single-token separator to the label, for example \verb|flowers: primrose| or \verb|coco: <the caption>|.
Although not strictly necessary, we mostly unify image preprocessing across all tasks: Inception-style cropping~\cite{szegedy15inception} with minimum area of 50\%, resize to 224\,px.
For classification tasks, we additionally randomly flip the image horizontally as is common practice; and for OCR-VQA~\cite{Mishra19ocrvqa}, we do not perform random cropping, as it resulted in very bad performance due to the nature of the task. 
We fix label smoothing to 0.1 and do not use randaugment, as these seemed to be good choices across all single-task setups.
Finally, we usually train for 50 epochs, which may not be enough to achieve maximum performance, but allows us to investigate more questions.

\section{Results and insights}\label{sec:results}

In this section we go through a series of results and insights regarding the use of autoregressive decoders for multi-tasking, each subsection aiming to answer one question or look at one aspect.
Unless specified, all experiments are performed using the same frozen pretrained encoder, in order to isolate any observed effects to the decoder itself.

\subsection{Single- vs multi-task}\label{sec:results:core}

We first answer the core question of how performance changes when going from single-task \lid models to using one multi-task \lid model for all the tasks.
For the single-task performance, we use the same autoregressive decoder but tune it heavily on each task separately, giving an indication of how good such a model architecture can perform on each task.
For the classification-style tasks, we also include a single-task feedforward baseline that is the best of a linear, an MLP, or an attention-based classifier.
For the multi-task \lid models we evaluate three different settings: a task-conditioned decoder, a category-conditioned decoder and an unconditioned decoder.
Results are summarized in Table~\ref{tbl:resuls:core}, where for each multi-task row, all numbers come from the same, single model.
It is clear that the unconditioned decoder's only chance of knowing what type of output is expected at test-time is by trying to identify what dataset an image may be from.
For datasets which have images from a specific domain such as RESISC45~\cite{cheng17resisc}, the unconditioned model performs at par with the task-conditioned setting.
In other cases, the model defaults to ImageNet as it is the largest dataset in the mix, which is evident in the poor performance of captioning tasks.
By at least telling the decoder what \emph{kind} of output is expected (category-conditioned), it has a much higher success at solving the individual tasks, although it cannot exploit individual dataset's peculiarities, which might be a good thing. As expected, captioning performance improves significantly in the category-conditioned setting, but performance for Flickr30k is still lower than for the task-conditioned model as COCO is roughly 5 times larger.
Finally, by telling the decoder exactly which task it is expected to solve (multi-task \lid), it can produce an answer that fits exactly that type of task. We investigate this in more detail in Section~\ref{sec:results:cond_method}.
Overall, with task-conditioning, a single decoder model is able to perform mostly on-par with individual single-task decoders.

\subsection{Scaling number of tasks and decoder depth}\label{sec:results:scaling_tasks}

\begin{figure}
    \centering
    \includegraphics[width=1.0\linewidth]{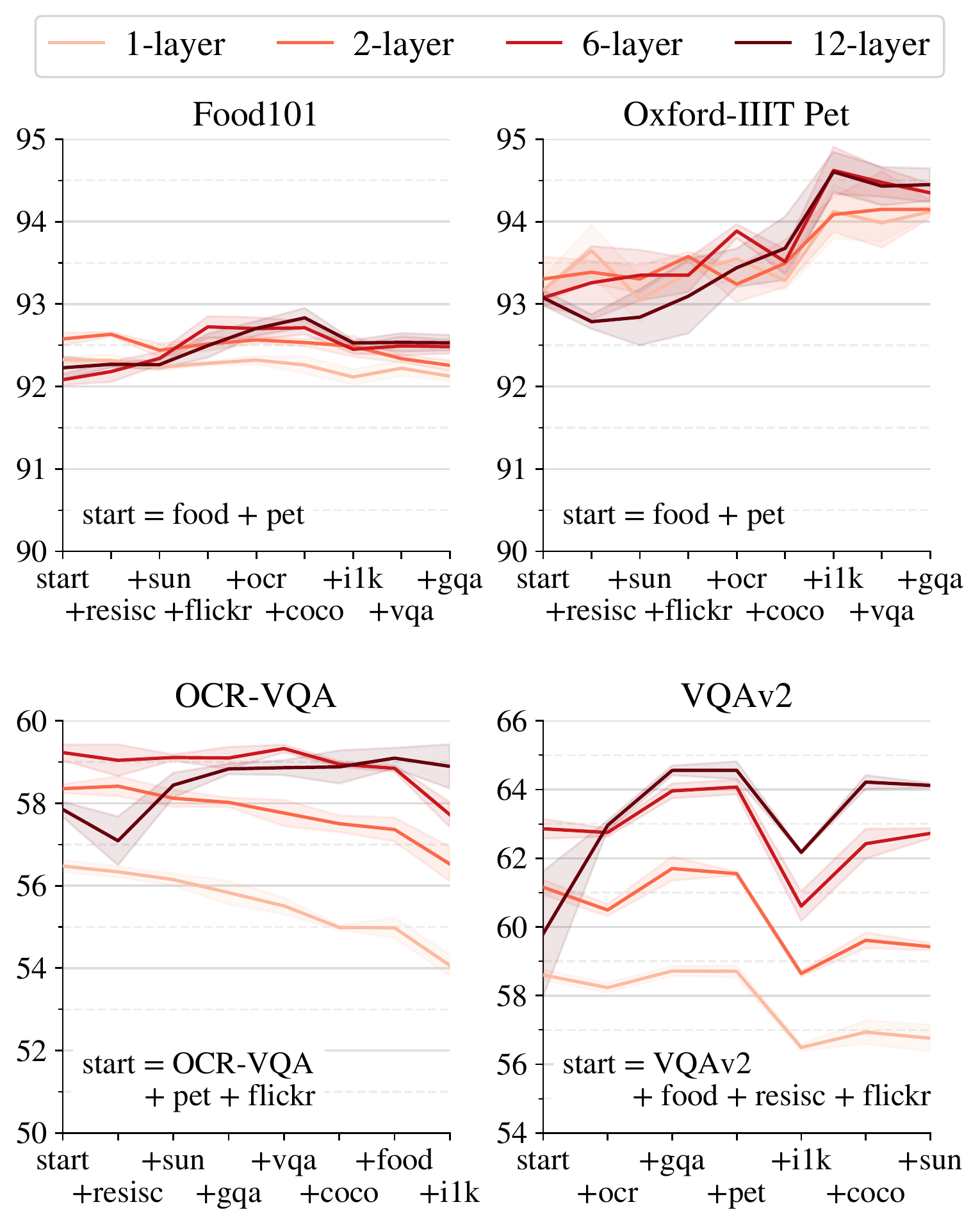}
    \caption{The effect of scaling the number of tasks and the decoder depth.
    Tasks are added from left to right and performance on one specific task is traced.
    Each point on the curves is the average performance (and stddev) of fully training from scratch three times.
    The top two plots show that there is no negative effect on classification-style tasks when increasing the load, for any decoder size.
    The bottom two plots, however, show that for more textual tasks, the performance of smaller decoders deteriorates with the addition of tasks. See text for detailed discussion.}
    \label{fig:ntask_vs_nlayer}
\end{figure}

The results presented so far were obtained using a 12-layer Base decoder
in order to avoid bottlenecks due to capacity.
An important question, however, is whether and how the capacity of the decoder needs to grow with the number of tasks.
We study this on four tasks: two classification tasks, OCR-VQA, and VQAv2.
We measure how the performance changes on these tasks as we train models on multi-task mixtures of increasing size.
Figure~\ref{fig:ntask_vs_nlayer} depicts this.
Each point is a full, separate training run on the mixture including all the tasks to its left; hence, the rightmost 12-layer point corresponds to the score of the full model in Table~\ref{tbl:resuls:core}.

The results clearly show different behaviors for different types of task.
First, classification-style tasks seem to be largely unaffected by \emph{both} the number of tasks and the decoder depth.
A notable but intuitive exception is Pets getting a significant boost by the inclusion of ImageNet-1k, which indeed contains many pet classes.
For more textual tasks such as OCR-VQA or VQAv2, however, we observe that for smaller decoders, performance gradually degrades with increased load.
Larger decoder models are less affected, with the 12-layer decoder keeping mostly constant performance.
Again, there are two notable exceptions: (a) a dip when ImageNet-1k is added, likely because it contains more than half of the mixture's images while providing little relevant signal (see Section~\ref{sec:results:mixing} for more);
and (b) the 12-layer decoder overfits on the small starting set of tasks, mainly because it is prohibitive to re-tune every single combination in this experiment (see Section~\ref{sec:results:regularization} for more).

\subsection{Task mixing strategies}\label{sec:results:mixing}

\begin{figure}
    \centering
    \includegraphics[width=1.0\linewidth]{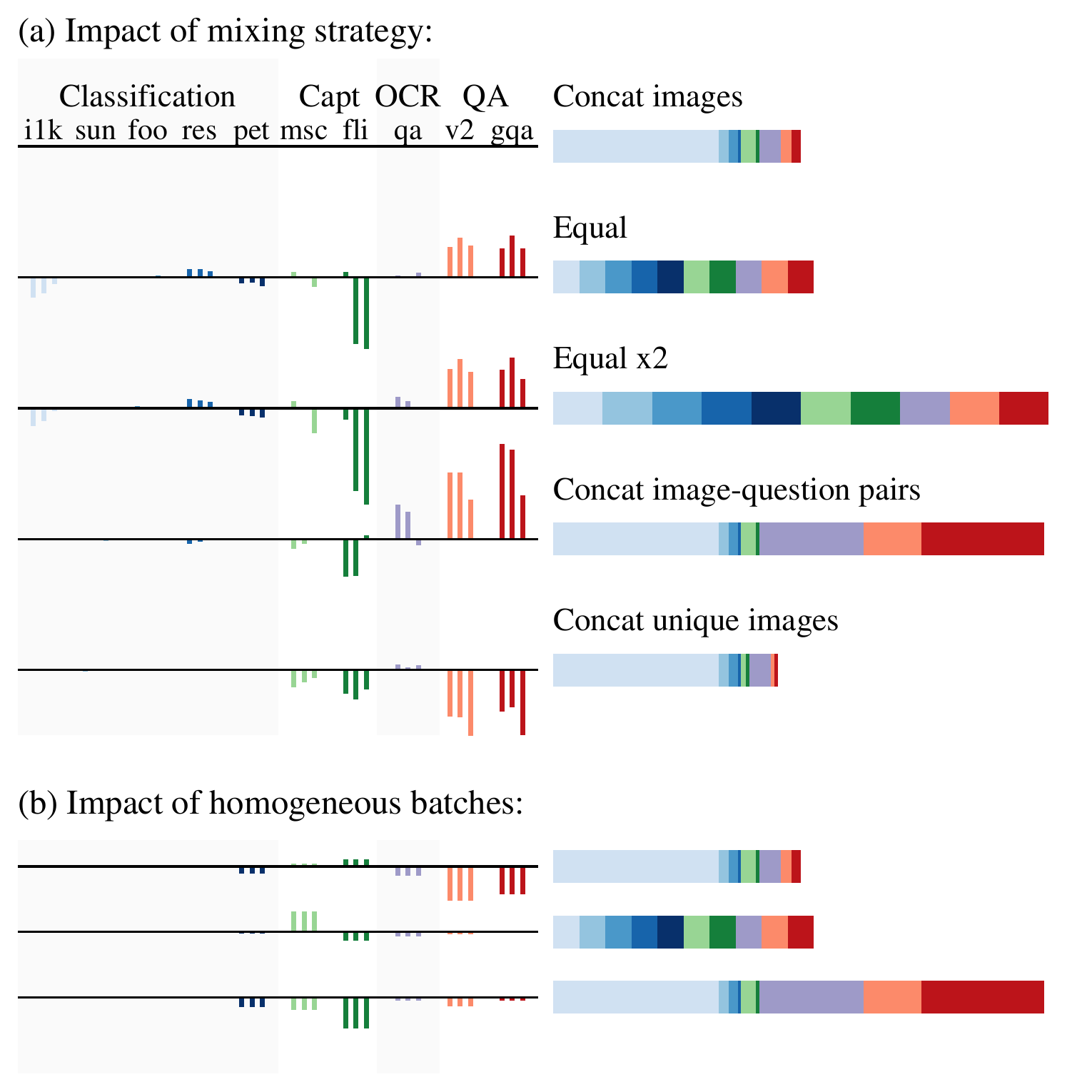}
    \caption{The effect of changing task mixing strategies.
    The bars on the right visualize the task proportions resulting from a few mixing strategies (see text).
    The bars on the left show the change in performance for each task when using a different strategy than \emph{Concat images}, for 1-layer, 2-layer, and 12-layer decoders.}
    \label{fig:mixing}
\end{figure}

An important decision when training on many tasks is how to mix them.
There are two distinctions to be made: the images, and the outputs.
A lot of multi-tasking work focuses on a single dataset, such as NYU Depth V2~\cite{Silberman12depth} or MS~COCO~\cite{lin14mscoco}, for which many different tasks and annotations are provided.
However, the question of mixing becomes even more important when the tasks use different image datasets.
We study both aspects simultaneously: our tasks span a wide range of datasets and dataset sizes (from 1.28\,M images in ImageNet to 3\,000 images in Pets) but also include three different tasks on MS~COCO (Captioning, VQAv2, GQA).
We propose a few mixing strategies:
\PAR{Concat images.} Our default strategy is concatenating all the images, regardless of their source and amount of annotations.
Each dataset is used for the same number of epochs, and more than half the examples come from ImageNet.
\PAR{Equal.} Sample from each task in equal proportion, and match the total training time of \emph{Concat images}. This leads to an extreme imbalance in terms of epochs on the datasets.
\PAR{Concat image--question pairs.} Like \emph{Concat images}, but each image--question pair is treated as one example. For instance, OCR-VQA contains on average 4.8 questions per image, and hence is visited 4.8 times more often.
\PAR{Concat unique images.} Like \emph{Concat images} except that images are de-duplicated, meaning that, for example, the three MS~COCO tasks are seen three times fewer.
\vskip4pt 

Figure~\ref{fig:mixing} shows the effect of each mixing strategy.
The main takeaway is that classification-style tasks are mostly unaffected, but besides that, mixing correctly is nontrivial.
However, visiting datasets with multiple labels per image more often seems beneficial, if one can control overfitting.

Another question is whether to mix different tasks within the same batch, or alternate between tasks across batches.
From a design standpoint, each have their separate pros and cons.
Mixing them within the batch seems to perform slightly better (see Figure~\ref{fig:mixing}), likely due to less biased gradients in each step.
However, using homogeneous batches would allow to better leverage the Transformer's flexibility: one could use different encoders or sequence lengths for the different tasks. We leave further exploring this aspect as promising future work.

\subsection{On transferring skills across tasks}\label{sec:results:transfer}

Here, we aim to see how much a skill learned in one task can help other, potentially lower-resource tasks in the mixture, or whether tasks are completely isolated after conditioning.
Our surprising finding, which is consistent across four very different settings, is that \emph{there is no apparent transfer of ``skills,'' merely the addition of meaningful tasks has a regularizing effect improving the decoder's performance on unrelated tasks}.
We are not aware of any experiments in the literature that are controlled enough to allow making this conclusion.
It is simultaneously a good and a bad news for multi-task learning.

\subsubsection{Auxiliary OCR tasks}
\input{tbls/results_ocr}
Our first experiment focuses on OCR.
We start from training on OCR-VQA alone, and then add an auxiliary task that is aimed to help learning to perform OCR.
The auxiliary task uses a 100\,M image subset of WebLI~\cite{chen22pali} where each image contains 2 to 20 OCR annotations obtained via the GCP Vision API.
We formulate three different OCR-related auxiliary tasks:
\PAR{OCR Concat} uses the concatenation of all OCR strings, from top-left to bottom-right, as labels.
\PAR{OCR Random} chooses one OCR string at random as label.
\PAR{OCR SplitCap}~\cite{chen22pali} is like WebLI-OCR Random but prompts the decoder with the first token of the label to indicate which text should be read.
\PAR{Alt-Text} adds WebLI images \emph{that do not have any detected text} and uses the image's alt-text as label. This controls for any improvement coming from auxiliary OCR tasks.
\vskip4pt 

Table~\ref{tbl:resuls:ocr} shows that the boost coming from auxiliary OCR tasks is at best in the same ballpark as that coming from this control experiment, leading to the same conclusion as in the previous subsection.

\subsubsection{Class tokens vs. class strings}
In our second experiment, we add a special token for each classification label over all datasets and always tokenize class labels into only one of the added tokens. If we would measure a significant drop in performance over tasks, that could indicate knowledge transfer due to sharing tokens over different tasks. In Figure~\ref{fig:results:tokenclass}, we observe very slight changes in performance, which means that there's no significant transfer due to token sharing.

\subsubsection{Treating languages as tasks}\label{sec:results:langs}

\begin{figure}[t]
\centering
\begin{minipage}[t]{.48\linewidth}
  \includegraphics[width=1.0\columnwidth]{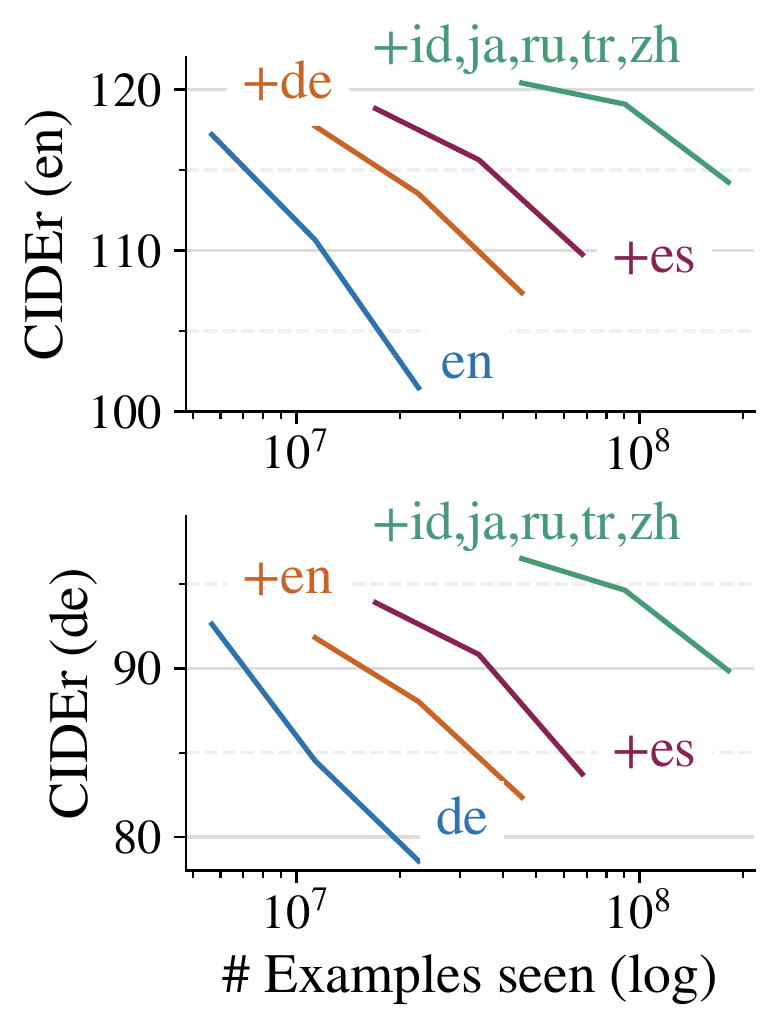}
  \caption{Adding more languages to the task mixture, even those with their own script and tokens, reduces overfitting and consistently improves performance, without transfer effect.}\label{fig:results:coco35l}
\end{minipage}\hfill
\begin{minipage}[t]{.48\linewidth}
  \includegraphics[width=1.1\columnwidth]{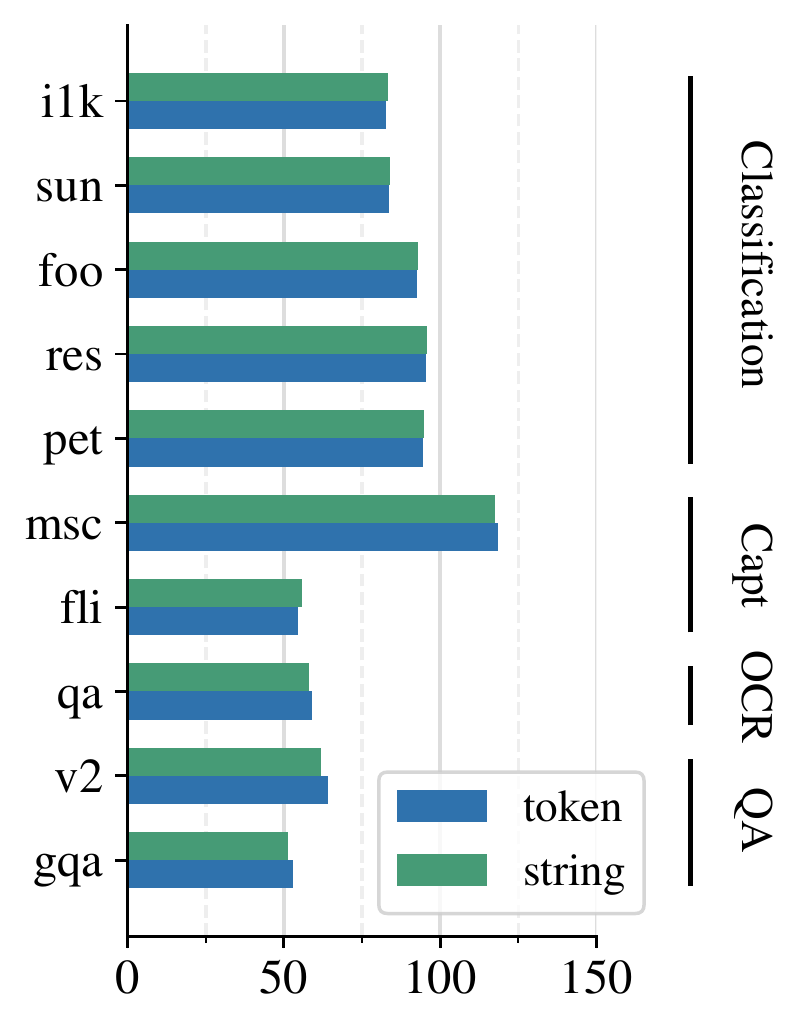}
  \caption{Replacing class names by unique, separate, meaningless class tokens makes no significant difference, indicating there is no knowledge transfer from class labels.}\label{fig:results:tokenclass}
\end{minipage}\hfill
\end{figure}

In a third experiment, we use machine-translated~\cite{thapliyal22crossmodal} COCO captions~\cite{chen15coco} as auxiliary tasks with varying degrees of label overlap, and identical images.
We treat each language as a separate task by prefixing its language code.
We tune every combination separately.
Figure~\ref{fig:results:coco35l} shows how the performance on English and German evolves as we see more images, either by visiting one setting for more epochs or by including more languages.
The results reveal that training on more languages improves performance by reducing overfitting.
While for related Latin languages this might be intuitive, it holds even in the case of adding languages with a non-Latin script.
Hence, this experiment provides further evidence toward the same conclusion.

\subsubsection{Multi-tasking reduces need for regularization}\label{sec:results:regularization} 
\begin{figure}
    \centering
    \includegraphics[width=1.0\linewidth]{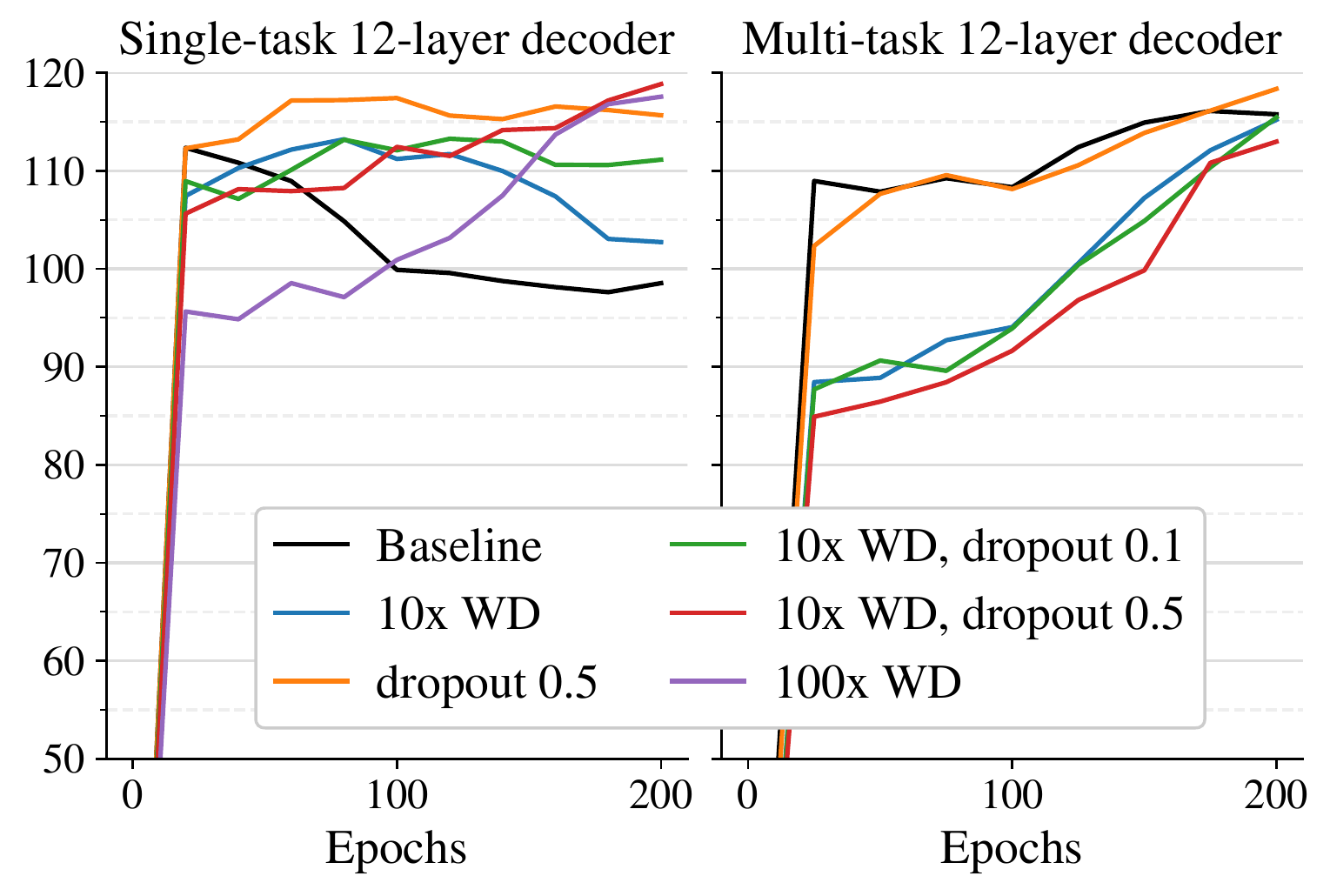}
    \caption{The multi-task setup is much less sensitive to regularization hyper-parameters than the single-task setup.
    Especially when training for very long, the single-task performance varies dramatically for different settings and shows the need for heavy tuning, whereas the multi-task setting reaches similar final performance across the board.}
    \label{fig:decoder_overfit}
\end{figure}

We noticed, and quanitify in this fourth experiment, that multi-tasking seems to reduce the need for carefully tuning regularization hyperparameters.
COCO captioning is sensitive to regularization-related hyperparameters such as weight-decay (WD) and dropout.
It is common to run pretrained models on COCO captioning for 5 or 10 epochs~\cite{li-etal-2022-mplug,wang22ofa,li2022blip2,hu2022expansionnet} and up to at most 40 epochs~\cite{hu22lemon}.
Figure~\ref{fig:decoder_overfit} shows COCO CIDEr evaluation curves for a large decoder intentionally trained for too long.
In the single-task case, the default setup (learning-rate 0.001, WD 1e-4, dropout 0.1) heavily overfits and the regularization hyper-parameters need to be tuned just right in order to get good results.
In the multi-task case, however, all of these settings reach similar final performance, thus the multi-task setup benefits from reduced need of hyper-parameter tuning.

\subsection{Importance of image encoder pretraining}\label{sec:results:nonwebli}

\begin{figure}[b]
    \centering
    \includegraphics[width=1.0\linewidth]{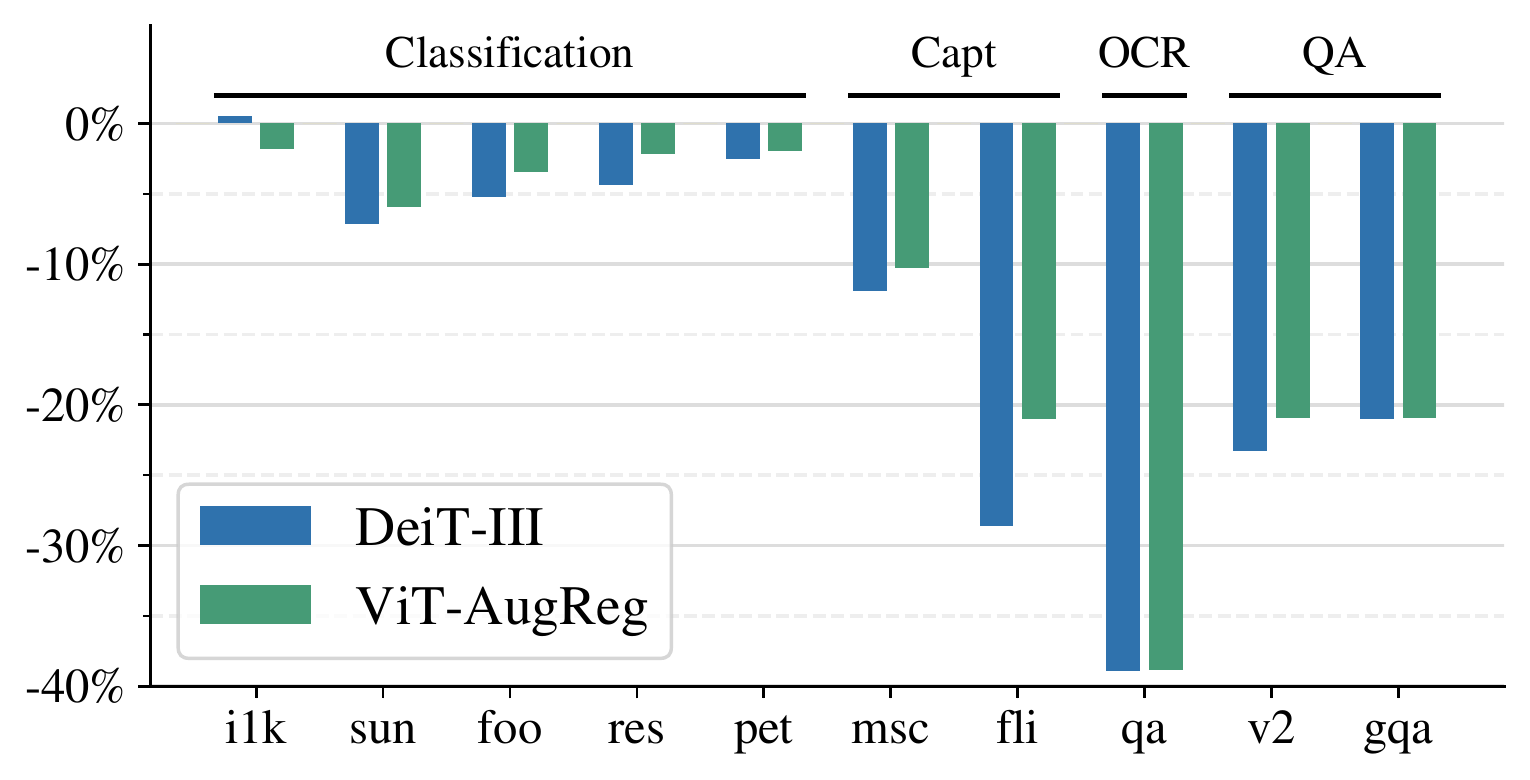}
    \caption{Relative changes of individual task metrics when training a frozen vision encoder pretrained on ImageNet-21k~\cite{Touvron2022DeiTIR,steiner22how} in the multi-task setup.
    Same data in table format in the appendix.
    }
    \label{fig:results:i21k}
\end{figure}

So far, we have used a frozen image encoder pretrained contrastively on a broad collection of images from the web.
Intuitively, such pretraining should result in a more general encoder than the predominant paradigm of pretraining on a large classification dataset~\cite{kolesnikov2020bit,Dosovitskiy21AnII,tan21efficnet2,steiner22how,liu2022convnet,Touvron2022DeiTIR}.
Hence, we also evaluate two models of the exact same architecture, trained on the broadest publicly available classification dataset ImageNet-21k~\cite{Olga15ILSVRC}: DeiT-III~\cite{Touvron2022DeiTIR} and ViT-AugReg~\cite{steiner22how}.
Sticking to the same architecture isolates the effect to the pretraining data and method.

The results in Figure~\ref{fig:results:i21k} show that the image encoder pretrained on image/text pairs performs significantly better in all tasks (other than INet-1k, where the DeiT-III backbone, which has been finetuned on ImageNet-1ik, performs on par). The difference is moderate with classification tasks, and larger with captioning, OCR and QA tasks.

\subsection{Task conditioning with prompt}\label{sec:results:cond_method}

Using a single decoder for multiple task adds the complexity that the model has to identify the task being solved and adjust its output distribution accordingly. As seen in Section \ref{sec:results:core}, an unconditional decoder may be able to perform this to some extent by inspecting the image. Adding a task prompt such as \verb|coco:|, to which the decoder can self-attend is critical to improve the model performance. In this section we further study this conditioning mechanism.

\begin{figure}[t]
    \centering
    \includegraphics[width=1.0\linewidth]{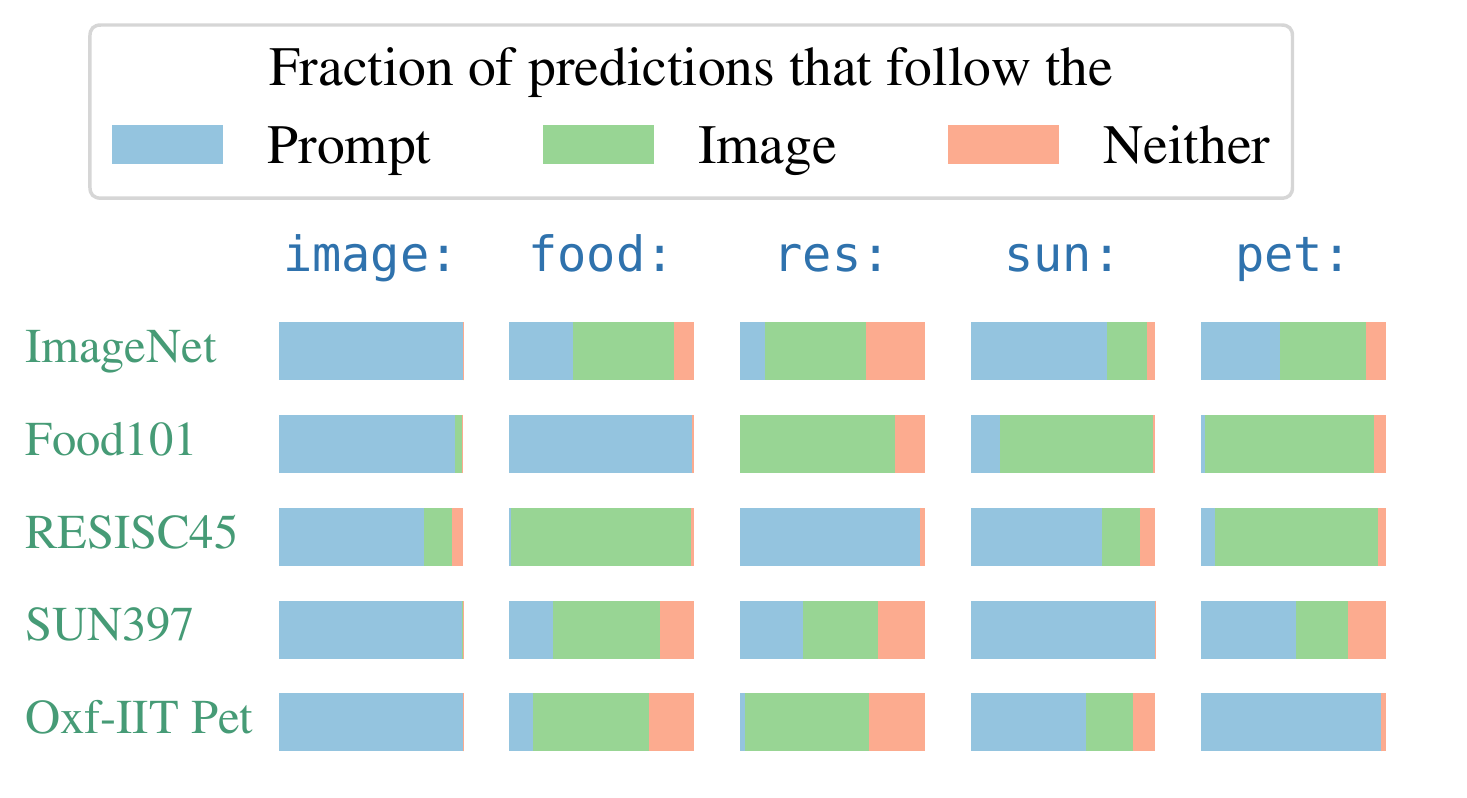}
    \caption{Distribution of output classes for various classification prompts (columns) on each dataset's validation images (rows).}
    \label{fig:results:prompts}
\end{figure}

We examined the effect of prompting the multi-task \lid with various prompts matching or not matching the image.
The results shown in Figure~\ref{fig:results:prompts} indicate that the model has a clear preference for labels corresponding to prompt and image, except for the case when prompt and image are highly mismatched (\eg \verb|res:| prompt on ImageNet dataset).
Additionally, for the \verb|image:| prompt, the predictions lie largely within the domain of ImageNet classes due to larger representation in the training mixture and due to the broad variety of images in ImageNet dataset. However for images that are completely out of domain for this prompt (\eg RESISC45), the model tends to predict classes from the image domain more often.  
These results indicate that conditioning with a prompt is not enough to force the model to operate in a specific task, in particular when the image is out of distribution.

Additionally, since we explore shallow decoders with 1 and 2 layers, we question whether making the information about the task being solved more readily available could help. To that end, we attempt an alternative task-conditioning method.
Instead of prompts, we use additional and distinct position embeddings for each task, which allows the model to earlier-on perform different computations depending on the task being solved. We run this 3 times using the hyper-parameters selected for the prompt setting for 1 and 2 layers setup and observed no significant difference.

\subsection{Frozen vs finetune image encoder}\label{sec:results:nonfrozen}

So far we have concentrated on \lid, which learns an autoregressive decoder on top of a frozen, pretrained encoder to allow for more controlled experiments.
In this section we run the experiments from 
Section~\ref{sec:results:core} in two further modes:
finetuning the pretrained encoder and training the encoder from scratch.

The results in the appendix show that finetuning the encoder also works equally well, but with additional finetuning cost compared to freezing the encoder.
Not surprisingly, in the from-scratch multi-task learning setup, all of the results go down significantly due to randomly initialized encoder weights. 
However, multi-task learning shows a larger gain compared to the above results using a pretrained encoder. 
This is expected as pooling together examples from different datasets allow the randomly initialized model to be trained on more data points.
Overall, our conclusions stay the same as the above study performed on a frozen encoder. 
And we recommend \lid with a frozen encoder to save compute, unless one observes a clear quality drop on a specific task.
We will expand the discussion in Section~\ref{sec:results:dense}.

\subsection{Using autoregressive decoders at test-time}\label{sec:results:decoding}

\begin{figure}
    \centering
    \includegraphics[width=1.0\linewidth]{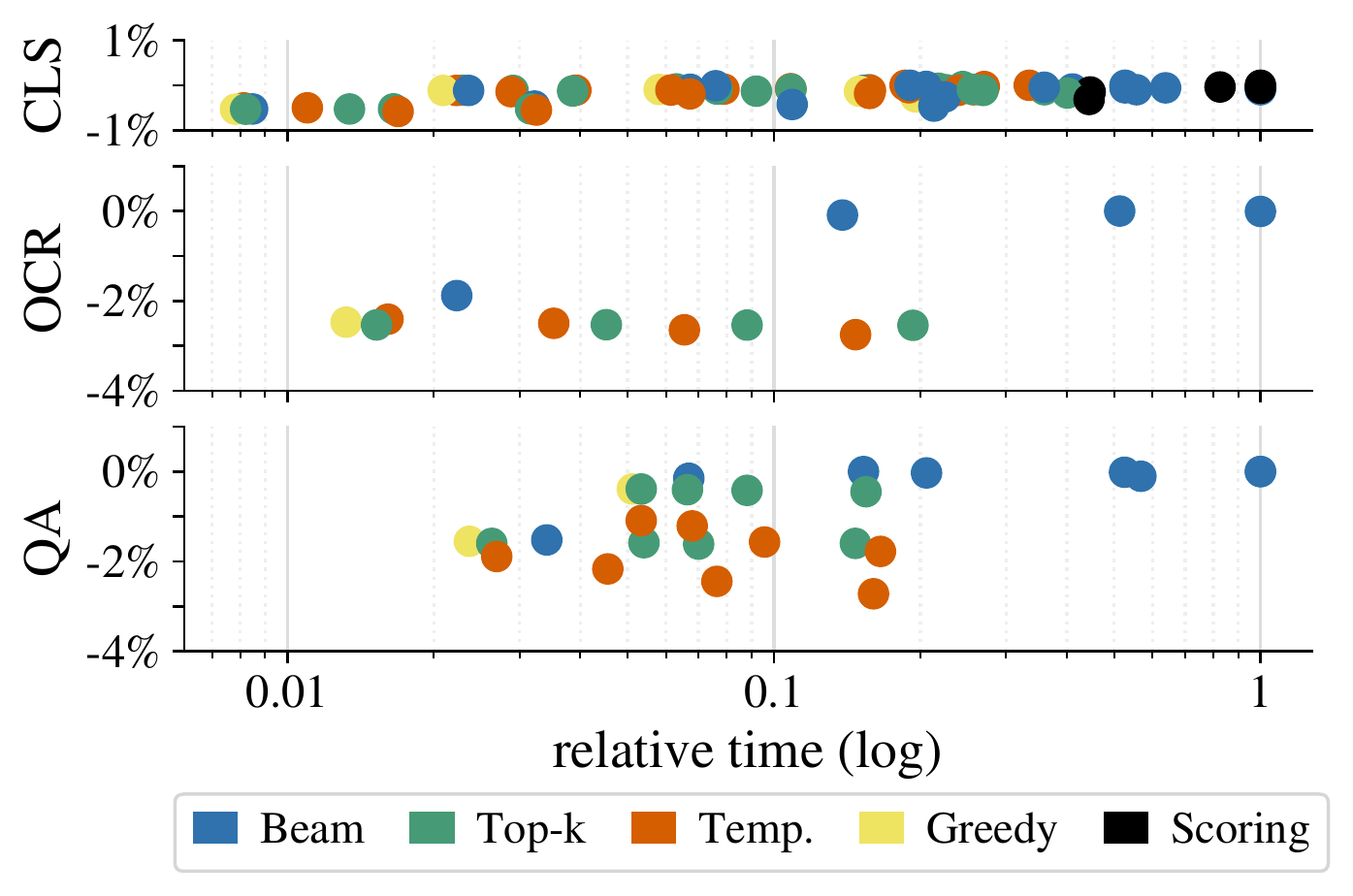}
    \caption{
    Relative performance plotted against evaluation time for different decoding strategies.
    Both dimensions are normalized per task (the best performing method is at $0\%$, and the slowest method at $1$).
    Beam search, top-k, and temperature sampling are evaluated at 1, 4, 8, and 16 samples/beams.
}
    \label{fig:results:decoding}
\end{figure}

Once the model is trained, different decoding strategies can be applied to generate output tokens incrementally (ranking every possible combination of output tokens by likelihood would be prohibitively expensive).
One such strategy is \emph{temperature sampling}, where every token is sampled from the output distribution of the model conditioned on the previously sampled tokens (setting T=0 results in the \emph{greedy} strategy).
We also examined \emph{beam search}, where the top $k$ candidates (``beams'') are kept and extended at every step (for a detailed explanation, see \eg~\cite{vijayakumar16diverse}).
We compared this to the more recent \emph{top-k} method~\cite{fan18topk,holtzman20degeneration}, a temperature sampling from the best $k$ candidates (and/or with a probability above a threshold).
For classification tasks, we also tried \emph{scoring} every possible class name and pick the one with the highest likelihood.

Figure~\ref{fig:results:decoding} shows the relative change in individual task metrics and generation time when applying different decoding strategies with the same multi-task LiT-Decoder model.
For classification, we found that different decoding strategies varied over more than two decades in terms of compute time (with scoring and beam search being the most expensive), but the performance was practically identical.
Other tasks, such as OCR and QA, can benefit from more expensive decoding schemes, allowing to trade additional compute for modestly improved performance.
For captioning, the trends are less clear.
All results reported in this paper use \emph{beam search} for captioning and OCR (with 4 beams, a Gumbel noise~\cite{kool19gumbel} of 0, and a $\alpha$=0.1 for length normalization~\cite{wu16nmt}), and \emph{greedy} for all other tasks.

\subsection{Attending to the encoder tokens}\label{sec:results:encatt}

One practical advantage of a frozen encoders, is that one can potentially replace storing an image by storing only its embeddings, which may be beneficial in terms of privacy, efficiency, and storage cost.
However, in the case of an encoder-decoder Transformer, the decoder typically attends into the \emph{sequence} of encoder outputs.
In our example, these are $14\times14$ tokens of dimension 768. Even when stored at half-precision, that is twice the size of the $224^2\,$px image.

\begin{figure}
    \centering
    \includegraphics[width=1.0\linewidth]{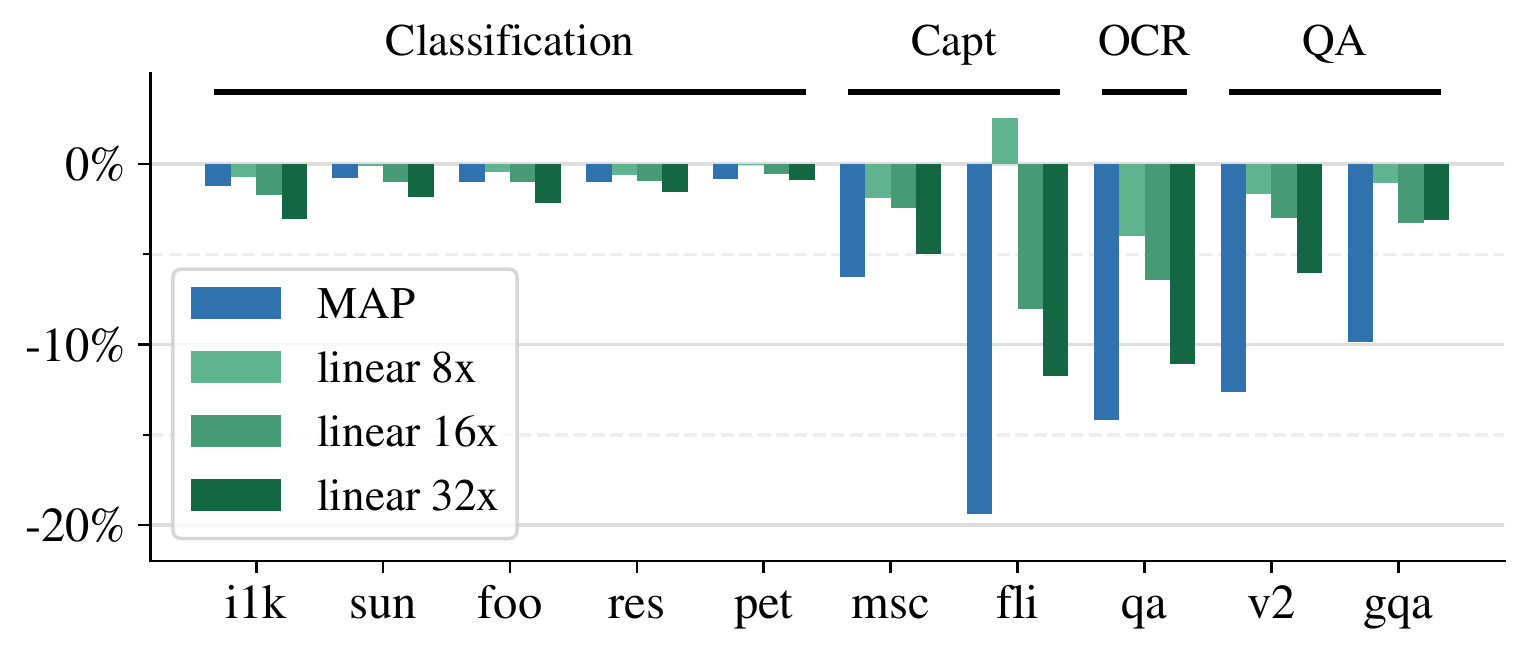}
    \caption{Summarizing the tokens into one via MAP, or reducing their size via a linear bottleneck (we tried 8x, 16x and 32x reduction) leads to a drop in quality across all tasks. The impact on classification tasks is small, but the other tasks degrade significantly the more we compress.}
    \label{fig:results:compression}
\end{figure}

We show in Figure \ref{fig:results:compression} that compressing the encoder output using several simple methods degrades the performance. An important new research direction for the future is to understand how to achieve better compression, without the significant performance drop. 

\subsection{Towards multi-tasking with dense targets}\label{sec:results:dense}

One limitation of the autoregressive decoder models is their scalability with respect to the output sequence length.
It makes it computationally prohibitive to directly tackle tasks with dense outputs like panoptic image segmentation which deals with an output in the order of million of pixels.

However, recent work~\cite{kolesnikov22uvim,lu23unifiedio} has demonstrated that targets of many dense vision tasks
can be compressed to relatively short discrete sequences.
We augment our study with the panoptic segmentation task, where mask targets are encoded using the VQVAE model~\cite{van2017neural} from UViM~\cite{kolesnikov22uvim}. This VQVAE maps a panoptic mask to a sequence of 256 values, where each element takes 4{,}096 distinct discrete values. 

For multi-task training with the panoptic segmentation task, we use Objects365~\cite{shao2019objects365}. We opted for this dataset because it is sufficiently large so we do not run into the overfitting issues. As the original dataset does not contain panoptic labels, we employ the state-of-the-art kMaX-DeepLab~\cite{yu2022k} to generate panoptic pseudo labels. The pretrained model is obtained from the open-source repository~\cite{weber2021deeplab2}. We evaluate the resulting model on the COCO panoptic dataset~\cite{chen15coco,kirillov2019panoptic} using the standard validation split.

This experiment yields an interesting insight: in contrast to our previous observations, the frozen pretrained image encoder does not work adequately for the panoptic image segmentation task
and achieves only 5.0\% PQ score. However, with finetuning setup we are able to obtain a large performance boost to 32.1\%, but in this case performance on other tasks slightly deteriorates (results in the appendix).
This indicates out-of-the-box pretrained vision encoders may not learn sufficiently good spatial semantic features. In the future we plan to further investigate and potentially address this shortcoming.

\section{Conclusion}\label{sec:conclusion}

In this paper, we aim to shed some light on the
training of multi-task computer-vision models with an autoregressive decoder.
For this, we mostly focus on a frozen pretrained encoder in order to study specifically if a single autoregressive decoder model is able to read-out the correct answer for each task from the image encoder.
This works surprisingly well, even with a small decoder, but only if the encoder was pretrained broadly enough, \eg CLIP-style with image--text data. 
However it still seems to fall short on dense output tasks.

Comparing to heavily-tuned single-task baselines, we find that when prompted for the task, a single multi-task decoder can match the baseline's performance.
In some cases, the decoder can even surpass it by benefiting from positive transfer despite the ``isolation'' of tasks via conditioning.
Overall, training a multi-task decoder seems less prone to overfitting and hence less sensitive to regularization hyper-parameters.
This is still true when regarding different languages as different tasks, painting an overall promising future for multi-task decoders.

\section{Acknowledgements}\label{sec:acks}
We thank Daniel Keysers for feedback on a draft of the paper.
We also thank Ilya Tolstikhin and Maxim Neumann for early implementations of autoregressive decoding strategies and beam search.
We use the {\tt big\_vision} codebase~\cite{big_vision1,big_vision2} for all experiments in this project.
As always, we thank the Google Brain team at large for providing a supportive research environment. 

{\small
\bibliographystyle{ieee_fullname}
\bibliography{egbib}
}

\clearpage

\input{appendix}

\end{document}

%% file: tbls/results_core.tex
\begin{table*}[t]
  \setlength{\tabcolsep}{0pt}
  \setlength{\extrarowheight}{5pt}
  \renewcommand{\arraystretch}{0.75}
  \newcolumntype{C}{>{\centering\arraybackslash}X}
  \newcolumntype{R}{>{\raggedleft\arraybackslash}X}
  \centering
  \caption{
    Results of using 12-layer auto-regressive decoders in single and multi-task (MT) settings.
    We report mean and standard deviation of 3 runs, bold indicates the two standard-deviation interval overlaps with the two standard-deviation interval of the best result. 
    We observe that a single decoder conditioned with a task (via prompt) and on a frozen image representation (via cross attention) can be trained to perform multiple computer vision tasks equally well as training a separate decoder for each task. The main differences observed are in INet-1k and Pet (discussed in main text).
    Additionally it shows conditioning on the task outperforms using task-type or no task conditioning.
  }\label{tbl:resuls:core}
\begin{tabularx}{\linewidth}{p{4cm}p{0.01cm}Cp{0.01cm}Cp{0.01cm}Cp{0.01cm}Cp{0.01cm}Cp{0.1cm}Cp{0.01cm}Cp{0.1cm}Cp{0.1cm}Cp{0.01cm}C}
  \toprule[1pt]
 && \multicolumn{9}{c}{Classification} && \multicolumn{3}{c}{Captioning} && OCR && \multicolumn{3}{c}{Question Ans.}\\
  \cmidrule[0.5pt]{3-11} \cmidrule[0.5pt]{13-15} \cmidrule[0.5pt]{17-17} \cmidrule[0.5pt]{19-21}

 && INet-1k && SUN397 && Food101 && RESISC45 && Pet && COCO && Flickr30k && -VQA && VQAv2 && GQA\\
  \midrule
Single-task \lid && \textbf{83.7{\scriptsize$\pm$0.0}} && \textbf{83.6{\scriptsize$\pm$0.1}} && \textbf{92.7{\scriptsize$\pm$0.0}} && \textbf{95.8{\scriptsize$\pm$0.1}} && 92.8{\scriptsize$\pm$0.2} && \textbf{118.8{\scriptsize$\pm$0.5}} && \textbf{50.0{\scriptsize$\pm$1.6}} && \textbf{60.1{\scriptsize$\pm$0.2}} && 63.6{\scriptsize$\pm$0.0} && \textbf{52.3{\scriptsize$\pm$0.3}} \\
Multi-task \lid && 82.8{\scriptsize$\pm$0.0} && \textbf{83.7{\scriptsize$\pm$0.1}} && \textbf{92.5{\scriptsize$\pm$0.1}} && \textbf{95.5{\scriptsize$\pm$0.1}} && \textbf{94.4{\scriptsize$\pm$0.2}} && \textbf{118.7{\scriptsize$\pm$0.5}} && \textbf{54.6{\scriptsize$\pm$0.8}} && \textbf{58.9{\scriptsize$\pm$0.4}} && \textbf{64.1{\scriptsize$\pm$0.1}} && \textbf{52.8{\scriptsize$\pm$0.2}} \\
\arrayrulecolor{lightgray}\midrule[0.25pt]\arrayrulecolor{black}
Category-conditioned MT && 82.6{\scriptsize$\pm$0.0} && 82.0{\scriptsize$\pm$0.0} && 92.2{\scriptsize$\pm$0.1} && 95.4{\scriptsize$\pm$0.0} && 75.0{\scriptsize$\pm$0.4} && \textbf{118.3{\scriptsize$\pm$0.7}} && 48.3{\scriptsize$\pm$1.4} && \textbf{59.9{\scriptsize$\pm$0.0}} && \textbf{64.2{\scriptsize$\pm$0.3}} && \textbf{52.1{\scriptsize$\pm$0.1}} \\
Unconditioned MT && 81.2{\scriptsize$\pm$0.1} && 81.7{\scriptsize$\pm$0.1} && 92.0{\scriptsize$\pm$0.0} && 95.2{\scriptsize$\pm$0.1} && 74.2{\scriptsize$\pm$0.3} && 50.3{\scriptsize$\pm$0.3} && 18.0{\scriptsize$\pm$0.3} && 57.1{\scriptsize$\pm$0.4} && 63.1{\scriptsize$\pm$0.2} && 51.5{\scriptsize$\pm$0.2} \\
\arrayrulecolor{lightgray}\midrule[0.25pt]\arrayrulecolor{black}
Single-task feedforward && 80.9{\scriptsize$\pm$0.1} && 83.3{\scriptsize$\pm$0.0} && 91.7{\scriptsize$\pm$0.0} && 93.4{\scriptsize$\pm$0.0} && 90.3{\scriptsize$\pm$0.2} && - && - && - && - && - \\
  \bottomrule
  \end{tabularx}
\end{table*}

%% file: tbls/results_ocr.tex
\begin{wraptable}{r}{4.1cm}
  \setlength{\tabcolsep}{0pt}
  \setlength{\extrarowheight}{5pt}
  \renewcommand{\arraystretch}{0.75}
  \newcolumntype{C}{>{\centering\arraybackslash}X}
  \newcolumntype{R}{>{\raggedleft\arraybackslash}X}
  \centering
  \caption{OCR skill, see text.}\label{tbl:resuls:ocr}
\begin{tabularx}{4.1cm}{p{2.8cm}p{0.01cm}C}
    \toprule[1pt]
OCR-VQA only && 54.8{\scriptsize$\pm$0.6} \\
    \arrayrulecolor{white}\midrule[0.25pt]\arrayrulecolor{black}
\multicolumn{3}{l}{\cellcolor{lightgray!10}\small{Added WebLI task:}} \\
OCR SplitCap && 53.5{\scriptsize$\pm$1.4} \\
OCR Random && 56.4{\scriptsize$\pm$0.6} \\
Alt-Text && 59.7{\scriptsize$\pm$0.1} \\
OCR Concat && 60.5{\scriptsize$\pm$0.1} \\
  \bottomrule
  \end{tabularx}
\end{wraptable}

%% file: appendix.tex
\appendix

\section{Datasets for multitask}

\input{tbls/task_list}

Table~\ref{tbl:task_list} gives an overview of the datasets we use across the paper, their absolute size, and origin.
Note that for VQAv2 our results are \emph{not} using the official test server, but only following the commonly used val-test split for which labels are available.

\section{Settings for single-task baselines}
\label{app:st_hp}

In order to get solid baseline numbers, we run a large sweep over hyper-parameters that we expect may have a significant impact on performance, select the best values according to performance on a validation set (either official, or held-out from the training set) and then re-run training three times using these best hyper-parameters.

\subsection{Feedforward classifiers}

We train a linear classifier by performing a full cross-product sweep over the following hyper-parameters for each of the 5 classification tasks:

\begin{itemize}
    \item Learning-rate: 0.001, 0.0003, 0.0001, 0.00003
    \item Weight-decay (decoupled): 0, 0.00001, 0.0001
    \item Number of epochs: 3, 10
    \item Pooling of encoder output:
    \begin{itemize}
        \item no pooling: embeddings are concatenated
        \item MAP frozen: weights of the Multi-head Attention Pooling (MAP) head~\cite{zhai22scaling} of the frozen backbone are reused for summarizing all embeddings into a single vector.
        \item MAP reset: a new MAP head is attached and trained from scratch.
    \end{itemize}
    \item Hidden layer dimension: no hidden layer (\ie linear classifier), and 1024 or 2048 dimensional unit (\ie MLP classifier).
\end{itemize}

Other notable settings (selected in smaller sweeps):
\begin{itemize}
    \item softmax cross-entropy loss performed better than sigmoid cross-entropy loss
    \item ScalingViT AdaFactor~\cite{zhai22scaling} performed better than SGD with momentum
    \item images were randomly cropped and flipped during training
    \item gradient clipping norm is set to 1.0
\end{itemize}

\subsection{Single-task decoder}

We perform a full cross-product sweep over the following hyper-parameters:

\begin{itemize}
    \item Learning-rate: 0.003, 0.001, 0.0003, 0.0001
    \item Weight-decay (decoupled): 0.01, 0.003, 0.001, 0.0003, 0.0001
    \item Decoder dropout: 0.1, 0.2, 0.5
    \item Image crops: resize only, random crop, inception\_crop with 50\% minimum area.
    \item Label smoothing: 0.0, 0.1
\end{itemize}

We also found in early, unstructured experiments that the following things do not have clear effect on performance or are universally detrimental:

\begin{itemize}
    \item RandAugment~\cite{Cubuk20randaugment}: it hurt performance across the board whenever we tried it.
    \item Dropout in the encoder: it either had no effect or hurt whenever we tried it.
    \item Random horizontal flips had no effect or hurt on any non-classification tasks.
    \item Adam/AdaFactor's $\beta_2$: No noticeable difference between 0.999 and 0.95.
\end{itemize}

\section{Settings for multi-task}
\label{app:mt_hp}

Because multi-task training runs for longer, we did not perform as extensive hyper-parameter search.
For some less important hyper-parameters, we stick to what seemed to work best across most single-task runs: inception\_crop, label smoothing 0.1.
We initially used the learning-rate, weight-decay, and decoder dropout that seemed to work reasonably well across all single-task decoder baselines, however that regularized too aggressively and the model clearly underfit.
Hence, we ran a few individual experiments with reducing each of those, and found that while not absolutely optimal, the ``common default'' values (used in many diverse publications) of learning-rate 0.001, (decoupled) weight-decay 0.0001, and dropout 0.1 worked quite well across tasks, so we decided to simply use these.

\section{Results for varying decoder depth}

Here, we present the same core result as in Table~\ref{tbl:resuls:core} but changing the decoder's size by reducing its depth to 1 or 2 layers in Tables~\ref{tbl:resuls:core:d1} and~\ref{tbl:resuls:core:d2}, respectively.

Two surprising results can be read from these tables.
First, a single-layer decoder is surprisingly good, with no drop in single-task performance on classification and captioning, and only a minor drop on more complicated OCR-VQA and QA tasks.
And second, while multi-tasking performance does slightly drop for this shallow models, it is close and consistent enough to suggest that one can prototype and explore with such extremely lightweight models, and then simply scale up.

\input{tbls/results_core_d1}
\input{tbls/results_core_d2}

\section{Results for various pretrained models}
\input{tbls/results_pretrained_models}

The results in Table~\ref{tbl:results:i21k} and Figure~\ref{fig:results:i21k} are based on three runs with different seeds and using the same learning rate (0.001) and weight decay (0.0001) that we used for the multi-task decoder in Table~\ref{tbl:resuls:core}. The image encoder was kept frozen during the multi-task training. The ViT-AugReg model from~\cite{steiner22how} was only pretrained on ImageNet-21k, and the DeiT-III model from~\cite{Touvron2022DeiTIR} was both pretrained on ImageNet-21k, and then finetuned on ImageNet-1k. This explains the slightly better ImageNet performance of the multi-task decoder using a DeiT-III backbone, at the cost of decreased performance in all the other tasks (compared to the less specialized ViT-AugReg backbone). For a discussion of the comparison with the multi-task baseline trained on a more diverse image-text dataset, refer to Section~\ref{sec:results:nonwebli}.

\section{Results for frozen/finetune/from-scratch image encoder}\label{app:finetune}

Table~\ref{tbl:resuls:fs_ft} shows single-task and multi-task results with frozen, finetune, and from-scratch image encoders.
We use similar hyper parameter sweep range as described in Appendix~\ref{app:st_hp} and Appendix~\ref{app:mt_hp} for finetune and from-scratch setups.
For fine-tune, we apply a 0.1 learning rate multiplier to the encoder weights by default.
The results are discussed in Section~\ref{sec:results:nonfrozen}.

\input{tbls/results_fs_ft}

\section{Results for panoptic segmentation task}\label{app:pan}

 In section~\ref{sec:results:dense} we introduce initial experiments on adding the dense panoptic segmentation task. In this section we present the complete numerical results. We compare against the multitask baseline from Table~\ref{tbl:resuls:core}. If the vision encoder is frozen, we observe very low performance for the panoptic segmentation task and roughly the same performance on all other tasks. With non-frozen vision encoder the strong finetuning setup (10x initial higher learning rate compared to the paper's setting), we get descent results on the panoptic task, but performance on other tasks deteriorates. With finetuning setup from Section~\ref{sec:results:nonfrozen} we observe the performance similar to the the ``frozen'' setup. Overall, we conclude that the out-of-the-box standard frozen pretrained vision encoders may not be sufficient for solving dense segmentation tasks.

\input{tbls/results_uvim}

%% file: tbls/task_list.tex
\begin{table}[h]
  \setlength{\tabcolsep}{0pt}
  \setlength{\extrarowheight}{5pt}
  \renewcommand{\arraystretch}{0.75}
  \newcolumntype{C}{>{\centering\arraybackslash}X}
  \newcolumntype{R}[1]{>{\raggedleft\let\newline\\\arraybackslash\hspace{0pt}}m{#1}}
  \centering
  \caption{Overview of the various tasks explored.}\label{tbl:task_list}
\begin{tabularx}{\linewidth}{p{2.2cm}p{2.3cm}R{1cm}C}
  \toprule[1pt]
  Type & Task  & Train & Testset \\
  \midrule
  \multirow{5}{*}{Classification} & INet-1k~\cite{Olga15ILSVRC} & $1.2$\,M & {\tt validation} \\
  & SUN397~\cite{zhou2017places} & $76$\,K & {\tt test} \\
  & Food101~\cite{bossard14food} & $74$\,K & {\tt validation} \\
  & RESISC45~\cite{cheng17resisc} & $19$\,K & {\tt train[80\%:]} \\
  & Pet~\cite{parkhi12a} & $2~944$ & {\tt test} \\
  \midrule
  \multirow{2}{*}{Captioning} & COCO~\cite{chen15coco} & $112$\,K  & {\tt val} \\
  & Flickr30k~\cite{young14flickr} & $28$\,K & {\tt test} \\
  \midrule
  \multirow{1}{*}{OCR} & OCR-VQA~\cite{chen22pali} & $166$\,K & {\tt test} \\
  \midrule
  \multirow{2}{*}{\shortstack[l]{Question\\ Answering}} & VQAv2~\cite{Goyal17vqav2} & $83$\,K & {\tt val-test} \\
  & GQA~\cite{hudson19gqa} & $72$\,K & {\tt test\_dev} \\
  \bottomrule
\end{tabularx}
\end{table}

%% file: tbls/results_core_d1.tex
\begin{table*}[t]
  \setlength{\tabcolsep}{0pt}
  \setlength{\extrarowheight}{5pt}
  \renewcommand{\arraystretch}{0.75}
  \newcolumntype{C}{>{\centering\arraybackslash}X}
  \newcolumntype{R}{>{\raggedleft\arraybackslash}X}
  \centering
  \caption{
    Results of using \textbf{1-layer} auto-regressive decoders in single and multi-task (MT) settings.
    We report mean and standard deviation of 3 runs, bold indicates the two standard-deviation interval overlaps with the two standard-deviation interval of the best result.
  }\label{tbl:resuls:core:d1}
\begin{tabularx}{\linewidth}{p{4cm}p{0.01cm}Cp{0.01cm}Cp{0.01cm}Cp{0.01cm}Cp{0.01cm}Cp{0.1cm}Cp{0.01cm}Cp{0.1cm}Cp{0.1cm}Cp{0.01cm}C}
  \toprule[1pt]
 && \multicolumn{9}{c}{Classification} && \multicolumn{3}{c}{Captioning} && OCR && \multicolumn{3}{c}{Question Ans.}\\
  \cmidrule[0.5pt]{3-11} \cmidrule[0.5pt]{13-15} \cmidrule[0.5pt]{17-17} \cmidrule[0.5pt]{19-21}

 && INet-1k && SUN397 && Food101 && RESISC45 && Pet && COCO && Flickr30k && -VQA && VQAv2 && GQA\\
  \midrule
Single-task \lid && \textbf{82.6{\scriptsize$\pm$0.1}} && \textbf{83.1{\scriptsize$\pm$0.1}} && \textbf{92.3{\scriptsize$\pm$0.0}} && \textbf{95.7{\scriptsize$\pm$0.2}} && 92.7{\scriptsize$\pm$0.0} && \textbf{119.0{\scriptsize$\pm$0.0}} && 46.7{\scriptsize$\pm$0.6} && \textbf{56.3{\scriptsize$\pm$0.2}} && \textbf{60.1{\scriptsize$\pm$0.1}} && \textbf{49.8{\scriptsize$\pm$0.2}} \\
Multi-task \lid && 82.2{\scriptsize$\pm$0.1} && \textbf{83.3{\scriptsize$\pm$0.2}} && 92.1{\scriptsize$\pm$0.0} && 95.0{\scriptsize$\pm$0.1} && \textbf{94.1{\scriptsize$\pm$0.0}} && 116.5{\scriptsize$\pm$0.1} && \textbf{50.4{\scriptsize$\pm$0.3}} && 54.1{\scriptsize$\pm$0.2} && 56.8{\scriptsize$\pm$0.3} && 46.3{\scriptsize$\pm$0.3} \\
\arrayrulecolor{lightgray}\midrule[0.25pt]\arrayrulecolor{black}
Category-conditioned MT && 82.0{\scriptsize$\pm$0.1} && 79.9{\scriptsize$\pm$0.1} && 91.2{\scriptsize$\pm$0.1} && 94.0{\scriptsize$\pm$0.2} && 75.0{\scriptsize$\pm$0.2} && 116.5{\scriptsize$\pm$0.6} && 43.5{\scriptsize$\pm$1.4} && 55.2{\scriptsize$\pm$0.0} && 56.9{\scriptsize$\pm$0.1} && 45.7{\scriptsize$\pm$0.2} \\
Unconditioned MT && 80.8{\scriptsize$\pm$0.1} && 79.8{\scriptsize$\pm$0.1} && 91.0{\scriptsize$\pm$0.0} && 94.3{\scriptsize$\pm$0.1} && 75.1{\scriptsize$\pm$0.1} && 35.7{\scriptsize$\pm$0.5} && 8.9{\scriptsize$\pm$0.6} && 51.9{\scriptsize$\pm$0.4} && 56.0{\scriptsize$\pm$0.0} && 45.0{\scriptsize$\pm$0.2} \\
\arrayrulecolor{lightgray}\midrule[0.25pt]\arrayrulecolor{black}
Single-task feedforward && 80.9{\scriptsize$\pm$0.1} && \textbf{83.3{\scriptsize$\pm$0.0}} && 91.7{\scriptsize$\pm$0.0} && 93.4{\scriptsize$\pm$0.0} && 90.3{\scriptsize$\pm$0.2} && - && - && - && - && - \\
  \bottomrule
  \end{tabularx}
\end{table*}

%% file: tbls/results_core_d2.tex
\begin{table*}[t]
  \setlength{\tabcolsep}{0pt}
  \setlength{\extrarowheight}{5pt}
  \renewcommand{\arraystretch}{0.75}
  \newcolumntype{C}{>{\centering\arraybackslash}X}
  \newcolumntype{R}{>{\raggedleft\arraybackslash}X}
  \centering
  \caption{
    Results of using \textbf{2-layer} auto-regressive decoders in single and multi-task (MT) settings.
    We report mean and standard deviation of 3 runs, bold indicates the two standard-deviation interval overlaps with the two standard-deviation interval of the best result.
  }\label{tbl:resuls:core:d2}
\begin{tabularx}{\linewidth}{p{4cm}p{0.01cm}Cp{0.01cm}Cp{0.01cm}Cp{0.01cm}Cp{0.01cm}Cp{0.1cm}Cp{0.01cm}Cp{0.1cm}Cp{0.1cm}Cp{0.01cm}C}
  \toprule[1pt]
 && \multicolumn{9}{c}{Classification} && \multicolumn{3}{c}{Captioning} && OCR && \multicolumn{3}{c}{Question Ans.}\\
  \cmidrule[0.5pt]{3-11} \cmidrule[0.5pt]{13-15} \cmidrule[0.5pt]{17-17} \cmidrule[0.5pt]{19-21}

 && INet-1k && SUN397 && Food101 && RESISC45 && Pet && COCO && Flickr30k && -VQA && VQAv2 && GQA\\
  \midrule
Single-task \lid && - && \textbf{83.6{\scriptsize$\pm$0.0}} && 91.9{\scriptsize$\pm$0.0} && \textbf{95.5{\scriptsize$\pm$0.2}} && 92.8{\scriptsize$\pm$0.0} && \textbf{119.2{\scriptsize$\pm$0.4}} && 48.7{\scriptsize$\pm$1.4} && \textbf{58.5{\scriptsize$\pm$0.1}} && \textbf{62.2{\scriptsize$\pm$0.2}} && \textbf{52.0{\scriptsize$\pm$0.4}} \\
Multi-task \lid && \textbf{82.7{\scriptsize$\pm$0.1}} && \textbf{83.5{\scriptsize$\pm$0.1}} && \textbf{92.3{\scriptsize$\pm$0.0}} && \textbf{95.2{\scriptsize$\pm$0.1}} && \textbf{94.1{\scriptsize$\pm$0.1}} && 117.4{\scriptsize$\pm$0.2} && \textbf{53.8{\scriptsize$\pm$0.3}} && 56.5{\scriptsize$\pm$0.3} && 59.4{\scriptsize$\pm$0.1} && 48.9{\scriptsize$\pm$0.2} \\
\arrayrulecolor{lightgray}\midrule[0.25pt]\arrayrulecolor{black}
Category-conditioned MT && \textbf{82.6{\scriptsize$\pm$0.0}} && 81.5{\scriptsize$\pm$0.1} && 91.9{\scriptsize$\pm$0.1} && \textbf{94.9{\scriptsize$\pm$0.1}} && 75.5{\scriptsize$\pm$0.2} && 116.9{\scriptsize$\pm$0.4} && 46.8{\scriptsize$\pm$1.8} && 57.6{\scriptsize$\pm$0.0} && 59.4{\scriptsize$\pm$0.1} && 48.7{\scriptsize$\pm$0.3} \\
Unconditioned MT && 81.4{\scriptsize$\pm$0.0} && 81.2{\scriptsize$\pm$0.1} && 91.5{\scriptsize$\pm$0.1} && \textbf{94.9{\scriptsize$\pm$0.1}} && 75.6{\scriptsize$\pm$0.2} && 42.5{\scriptsize$\pm$0.5} && 10.9{\scriptsize$\pm$0.4} && 54.1{\scriptsize$\pm$0.2} && 58.3{\scriptsize$\pm$0.1} && 47.6{\scriptsize$\pm$0.1} \\
\arrayrulecolor{lightgray}\midrule[0.25pt]\arrayrulecolor{black}
Single-task feedforward && 80.9{\scriptsize$\pm$0.1} && 83.3{\scriptsize$\pm$0.0} && 91.7{\scriptsize$\pm$0.0} && 93.4{\scriptsize$\pm$0.0} && 90.3{\scriptsize$\pm$0.2} && - && - && - && - && - \\
  \bottomrule
  \end{tabularx}
\end{table*}

%% file: tbls/results_pretrained_models.tex
\begin{table*}[t]
  \setlength{\tabcolsep}{0pt}
  \setlength{\extrarowheight}{5pt}
  \renewcommand{\arraystretch}{0.75}
  \newcolumntype{C}{>{\centering\arraybackslash}X}
  \newcolumntype{R}{>{\raggedleft\arraybackslash}X}
  \centering
  \caption{
    Comparing multi-task results using an image encoder trained on a large diverse image-text dataset (the default setup described in Section~\ref{sec:setup}) with backbones pre-trained on the publicly available ImageNet-21k from \cite{steiner22how,Touvron2022DeiTIR}. Same data as in Figure~\ref{fig:results:i21k}.
  }\label{tbl:results:i21k}
\begin{tabularx}{\linewidth}{p{3.5cm}p{0.01cm}Cp{0.01cm}Cp{0.01cm}Cp{0.01cm}Cp{0.01cm}Cp{0.1cm}Cp{0.01cm}Cp{0.1cm}Cp{0.1cm}Cp{0.01cm}C}
  \toprule[1pt]
 && \multicolumn{9}{c}{Classification} && \multicolumn{3}{c}{Captioning} && OCR && \multicolumn{3}{c}{Question Ans.}\\
  \cmidrule[0.5pt]{3-11} \cmidrule[0.5pt]{13-15} \cmidrule[0.5pt]{17-17} \cmidrule[0.5pt]{19-21}

 && INet-1k && SUN397 && Food101 && RESISC45 && Pet && COCO && Flickr30k && -VQA && VQAv2 && GQA\\
  \midrule

Image-Text &&           82.8{\scriptsize$\pm$0.0} &&  \textbf{83.7{\scriptsize$\pm$0.1}} &&  \textbf{92.5{\scriptsize$\pm$0.1}} &&  \textbf{95.5{\scriptsize$\pm$0.1}} &&  \textbf{94.4{\scriptsize$\pm$0.2}} &&  \textbf{118.7{\scriptsize$\pm$0.5}} &&  \textbf{54.6{\scriptsize$\pm$0.8}} &&  \textbf{58.9{\scriptsize$\pm$0.4}} &&  \textbf{64.1{\scriptsize$\pm$0.1}} &&  \textbf{52.8{\scriptsize$\pm$0.2}} \\
DeiT-III   &&  \textbf{83.2{\scriptsize$\pm$0.0}} &&           77.7{\scriptsize$\pm$0.1} &&           87.7{\scriptsize$\pm$0.4} &&           91.3{\scriptsize$\pm$0.1} &&           92.1{\scriptsize$\pm$0.4} &&           104.5{\scriptsize$\pm$0.5} &&           38.9{\scriptsize$\pm$0.2} &&           36.0{\scriptsize$\pm$0.2} &&           49.2{\scriptsize$\pm$0.4} &&           41.7{\scriptsize$\pm$0.3} \\
ViT-AugReg &&           81.3{\scriptsize$\pm$0.0} &&           78.7{\scriptsize$\pm$0.1} &&           89.3{\scriptsize$\pm$0.1} &&           93.5{\scriptsize$\pm$0.2} &&           92.6{\scriptsize$\pm$0.1} &&           106.4{\scriptsize$\pm$0.3} &&           43.1{\scriptsize$\pm$1.5} &&           36.0{\scriptsize$\pm$0.2} &&           50.7{\scriptsize$\pm$0.4} &&           41.7{\scriptsize$\pm$0.1} \\

  \bottomrule
  \end{tabularx}
\end{table*}

%% file: tbls/results_fs_ft.tex
\begin{table*}[t]
  \setlength{\tabcolsep}{0pt}
  \setlength{\extrarowheight}{5pt}
  \renewcommand{\arraystretch}{0.75}
  \newcolumntype{C}{>{\centering\arraybackslash}X}
  \newcolumntype{R}{>{\raggedleft\arraybackslash}X}
  \centering
  \caption{
    12 layer decoder results with frozen, fine-tune and from-scrach encoder setups.
  }\label{tbl:resuls:fs_ft}
\begin{tabularx}{\linewidth}{p{4cm}p{0.01cm}Cp{0.01cm}Cp{0.01cm}Cp{0.01cm}Cp{0.01cm}Cp{0.1cm}Cp{0.01cm}Cp{0.1cm}C}
  \toprule[1pt]
 && \multicolumn{9}{c}{Classification} && \multicolumn{3}{c}{Captioning} && OCR\\
  \cmidrule[0.5pt]{3-11} \cmidrule[0.5pt]{13-15} \cmidrule[0.5pt]{17-17} 

 && INet-1k && SUN397 && Food101 && RESISC45 && Pet && COCO && Flickr30k && -VQA \\
  \midrule
Frozen (single-task) && 83.7{\scriptsize$\pm$0.0} && 83.6{\scriptsize$\pm$0.1} && 92.7{\scriptsize$\pm$0.0} && 95.8{\scriptsize$\pm$0.1} && 92.8{\scriptsize$\pm$0.2} && 118.8{\scriptsize$\pm$0.5} && 50.0{\scriptsize$\pm$1.6} && 60.1{\scriptsize$\pm$0.2} \\
Frozen (multi-task) && 82.8{\scriptsize$\pm$0.0} && 83.7{\scriptsize$\pm$0.1} && 92.5{\scriptsize$\pm$0.1} && 95.5{\scriptsize$\pm$0.1} && 94.4{\scriptsize$\pm$0.2} && 118.7{\scriptsize$\pm$0.5} && 54.6{\scriptsize$\pm$0.8} && 58.9{\scriptsize$\pm$0.4}  \\
\arrayrulecolor{lightgray}\midrule[0.25pt]\arrayrulecolor{black}
Finetune (single-task) && 83.1 && 82.6 && 93.0 && 96.8 && 91.2 && 122.3 && 45.5 && 61.7 \\
Finetune (multi-task) && 83.2 && 83.1 && 92.6 && 96.7 && 95.0 && 119.4 && 54.7 && 59.6 \\
\arrayrulecolor{lightgray}\midrule[0.25pt]\arrayrulecolor{black}
From-scratch (single-task) && 69.4 && 39.4 && 53.8 && 79.4 && 4.9 && 68.2 && 18.8 && 38.8 \\
From-scratch (multi-task) && 70.8 && 70.6 && 83.0 && 94.2 && 90.4 && 96.2 && 40.1 && 40.5 \\
  \bottomrule
  \end{tabularx}
\end{table*}

%% file: tbls/results_uvim.tex
\begin{table*}[t]
  \setlength{\tabcolsep}{0pt}
  \setlength{\extrarowheight}{5pt}
  \renewcommand{\arraystretch}{0.75}
  \newcolumntype{C}{>{\centering\arraybackslash}X}
  \newcolumntype{R}{>{\raggedleft\arraybackslash}X}
  \centering
  \caption{
    This table shows the effect of adding the extra dense segmentation task: we add coco panoptic segmentation, \textbf{Pan} (which uses UViM codes as mask encoding). 
  }\label{tbl:resuls:uvim}
\begin{tabularx}{\linewidth}{p{1.7cm}p{0.01cm}Cp{0.01cm}Cp{0.01cm}Cp{0.01cm}Cp{0.01cm}Cp{0.1cm}Cp{0.01cm}Cp{0.1cm}Cp{0.1cm}Cp{0.1cm}Cp{0.01cm}C}
  \toprule[1pt]
 && \multicolumn{9}{c}{Classification} && \multicolumn{3}{c}{Captioning} && OCR && \multicolumn{3}{c}{Question Ans.}\\
  \cmidrule[0.5pt]{3-11} \cmidrule[0.5pt]{13-15} \cmidrule[0.5pt]{17-17} \cmidrule[0.5pt]{19-21}

 && INet-1k && SUN397 && Food101 && RESISC45 && Pet && COCO && Flickr30k && -VQA && VQAv2 && GQA && Pan\\
  \midrule
  Baseline && 82.8{\scriptsize$\pm$0.0} && 83.7{\scriptsize$\pm$0.1} && 92.5{\scriptsize$\pm$0.1} && 95.5{\scriptsize$\pm$0.1} && 94.4{\scriptsize$\pm$0.2} && 118.7{\scriptsize$\pm$0.5} && 54.6{\scriptsize$\pm$0.8} && 58.9{\scriptsize$\pm$0.4} && 64.1{\scriptsize$\pm$0.1} && 52.8{\scriptsize$\pm$0.2} && -- \\
   Frozen && 83.1 && 83.8 && 92.7 && 95.4 && 94.2 && 117.5 && 55.5 && 58.1 && 61.5 && 50.7 && 5.1 \\
   FT strong && 78.6 && 77.6 && 87.8 && 95.3 && 93.3 && 113.0 && 49.5 && 45.8 && 59.1 && 49.6 && 32.1 \\
   FT paper && 82.8 && 82.2 && 92.4 && 96.3 && 95.0 && 118.1 && 55.5 && 59.8 && 62.4 && 51.1 && 7.3 \\
  \bottomrule
  \end{tabularx}
\end{table*}

%% file: arxiv.bbl
\begin{thebibliography}{100}\itemsep=-1pt

\bibitem{agrawal23reassessing}
Aishwarya Agrawal, Ivana Kaji{\'c}, Emanuele Bugliarello, Elnaz Davoodi, Anita
  Gergely, Phil Blunsom, and Aida Nematzadeh.
\newblock Reassessing evaluation practices in visual question answering: {A}
  case study on out-of-distribution generalization.
\newblock In {\em Findings of the Association for Computational Linguistics:
  EACL 2023}. Association for Computational Linguistics, May 2023.

\bibitem{Alam21MEDIC}
Firoj Alam, Tanvirul Alam, Md.~Arid Hasan, Abul Hasnat, Muhammad Imran, and
  Ferda Ofli.
\newblock {MEDIC:} {A} multi-task learning dataset for disaster image
  classification.
\newblock {\em arXiv preprint arXiv:2108.12828}, 2021.

\bibitem{antol15vqa}
Stanislaw Antol, Aishwarya Agrawal, Jiasen Lu, Margaret Mitchell, Dhruv Batra,
  C.~Lawrence Zitnick, and Devi Parikh.
\newblock Vqa: Visual question answering.
\newblock In {\em Proceedings of the IEEE International Conference on Computer
  Vision (ICCV)}, December 2015.

\bibitem{big_vision2}
Lucas Beyer, Xiaohua Zhai, and Alexander Kolesnikov.
\newblock Better plain vit baselines for imagenet-1k, 2022.

\bibitem{big_vision1}
Lucas Beyer, Xiaohua Zhai, and Alexander Kolesnikov.
\newblock Big vision.
\newblock \url{https://github.com/google-research/big_vision}, 2022.

\bibitem{Bhattacharjee22mult}
Deblina Bhattacharjee, Tong Zhang, Sabine Süsstrunk, and Mathieu Salzmann.
\newblock {MulT: An End-to-End Multitask Learning Transformer}.
\newblock In {\em CVPR}, 2022.

\bibitem{bossard14food}
Lukas Bossard, Matthieu Guillaumin, and Luc Van~Gool.
\newblock Food-101 -- mining discriminative components with random forests.
\newblock In {\em ECCV}, 2014.

\bibitem{bugliarello21unmasked}
Emanuele Bugliarello, Ryan Cotterell, Naoaki Okazaki, and Desmond Elliott.
\newblock {Multimodal Pretraining Unmasked: {A} Meta-Analysis and a Unified
  Framework of Vision-and-Language {BERT}s}.
\newblock {\em Transactions of the Association for Computational Linguistics},
  9:978--994, 2021.

\bibitem{bugliarello22iglue}
Emanuele Bugliarello, Fangyu Liu, Jonas Pfeiffer, Siva Reddy, Desmond Elliott,
  Edoardo~Maria Ponti, and Ivan Vuli{\'c}.
\newblock {IGLUE}: A benchmark for transfer learning across modalities, tasks,
  and languages.
\newblock In {\em ICML}, 2022.

\bibitem{carion20detr}
Nicolas Carion, Francisco Massa, Gabriel Synnaeve, Nicolas Usunier, Alexander
  Kirillov, and Sergey Zagoruyko.
\newblock End-to-end object detection with transformers.
\newblock In {\em ECCV}, 2020.

\bibitem{caruana1998multitask}
Rich Caruana.
\newblock {\em Multitask learning}.
\newblock Springer, 1998.

\bibitem{chen22pix2seq}
Ting Chen, Saurabh Saxena, Lala Li, David~J. Fleet, and Geoffrey Hinton.
\newblock Pix2seq: A language modeling framework for object detection.
\newblock In {\em ICLR}, 2022.

\bibitem{chen2022unified}
Ting Chen, Saurabh Saxena, Lala Li, Tsung-Yi Lin, David~J. Fleet, and Geoffrey
  Hinton.
\newblock A unified sequence interface for vision tasks.
\newblock In Alice~H. Oh, Alekh Agarwal, Danielle Belgrave, and Kyunghyun Cho,
  editors, {\em Advances in Neural Information Processing Systems}, 2022.

\bibitem{chen22asss}
Wuyang Chen, Xianzhi Du, Fan Yang, Lucas Beyer, Xiaohua Zhai, Tsung-Yi Lin,
  Huizhong Chen, Jing Li, Xiaodan Song, Zhangyang Wang, editor="Avidan~Shai
  Zhou, Denny", Gabriel Brostow, Moustapha Ciss{\'e}, Giovanni~Maria Farinella,
  and Tal Hassner.
\newblock A simple single-scale vision transformer for object detection
  and instance segmentation.
\newblock In {\em ECCV}, 2022.

\bibitem{chen15coco}
Xinlei Chen, Hao Fang, Tsung-Yi Lin, Ramakrishna Vedantam, Saurabh Gupta, Piotr
  Doll{\'a}r, and C.~Lawrence Zitnick.
\newblock Microsoft {COCO Captions}: Data collection and evaluation server.
\newblock {\em arXiv preprint arXiv:1504.00325}, 2015.

\bibitem{chen22pali}
Xi Chen, Xiao Wang, Soravit Changpinyo, AJ Piergiovanni, Piotr Padlewski,
  Daniel Salz, Sebastian~Alexander Goodman, Adam Grycner, Basil Mustafa, Lucas
  Beyer, Alexander Kolesnikov, Joan Puigcerver, Nan Ding, Keran Rong, Hassan
  Akbari, Gaurav Mishra, Linting Xue, Ashish Thapliyal, James Bradbury,
  Weicheng Kuo, Mojtaba Seyedhosseini, Chao Jia, Burcu~Karagol Ayan, Carlos
  Riquelme, Andreas Steiner, Anelia Angelova, Xiaohua Zhai, Neil Houlsby, and
  Radu Soricut.
\newblock {PaLI}: A jointly-scaled multilingual language-image model.
\newblock {\em arXiv preprint arXiv:2209.06794}, 2022.

\bibitem{cheng17resisc}
Gong Cheng, Junwei Han, and Xiaoqiang Lu.
\newblock Remote sensing image scene classification: Benchmark and state of the
  art.
\newblock {\em arXiv preprint arXiv:1703.00121}, 2017.

\bibitem{Crawshaw20multitask}
Michael Crawshaw.
\newblock Multi-task learning with deep neural networks: {A} survey.
\newblock {\em arXiv preprint arXiv:2009.09796}, 2020.

\bibitem{cubuk19randaug}
Ekin~D. Cubuk, Barret Zoph, Jonathon Shlens, and Quoc~V. Le.
\newblock Randaugment: Practical automated data augmentation with a reduced
  search space.
\newblock {\em arXiv preprint arXiv:1909.13719}, 2019.

\bibitem{Cubuk20randaugment}
Ekin~D. Cubuk, Barret Zoph, Jonathon Shlens, and Quoc~V. Le.
\newblock Randaugment: Practical automated data augmentation with a reduced
  search space.
\newblock In {\em Proceedings of the IEEE/CVF Conference on Computer Vision and
  Pattern Recognition (CVPR) Workshops}, June 2020.

\bibitem{Dosovitskiy21AnII}
Alexey Dosovitskiy, Lucas Beyer, Alexander Kolesnikov, Dirk Weissenborn,
  Xiaohua Zhai, Thomas Unterthiner, Mostafa Dehghani, Matthias Minderer, Georg
  Heigold, Sylvain Gelly, Jakob Uszkoreit, and Neil Houlsby.
\newblock An image is worth 16x16 words: Transformers for image recognition at
  scale.
\newblock In {\em ICLR}, 2021.

\bibitem{fan18topk}
Angela Fan, Mike Lewis, and Yann~N. Dauphin.
\newblock Hierarchical neural story generation.
\newblock In Iryna Gurevych and Yusuke Miyao, editors, {\em Proceedings of the
  56th Annual Meeting of the Association for Computational Linguistics, {ACL}
  2018, Melbourne, Australia, July 15-20, 2018, Volume 1: Long Papers}, pages
  889--898. Association for Computational Linguistics, 2018.

\bibitem{girdhar2022omnivore}
Rohit Girdhar, Mannat Singh, Nikhila Ravi, Laurens van~der Maaten, Armand
  Joulin, and Ishan Misra.
\newblock {Omnivore: A Single Model for Many Visual Modalities}.
\newblock In {\em CVPR}, 2022.

\bibitem{Gong19Comparison}
Ting Gong, Tyler Lee, Cory Stephenson, Venkata Renduchintala, Suchismita Padhy,
  Anthony Ndirango, Gokce Keskin, and Oguz~H. Elibol.
\newblock A comparison of loss weighting strategies for multi task learning in
  deep neural networks.
\newblock {\em IEEE Access}, 2019.

\bibitem{Goyal17vqav2}
Yash Goyal, Tejas Khot, Douglas Summers{-}Stay, Dhruv Batra, and Devi Parikh.
\newblock Making the {V} in {VQA} matter: Elevating the role of image
  understanding in visual question answering.
\newblock In {\em CVPR}, 2017.

\bibitem{Heo2021rethinking}
Byeongho Heo, Sangdoo Yun, Dongyoon Han, Sanghyuk Chun, Junsuk Choe, and
  Seong~Joon Oh.
\newblock Rethinking spatial dimensions of vision transformers.
\newblock In {\em ICCV}, 2021.

\bibitem{holtzman20degeneration}
Ari Holtzman, Jan Buys, Li Du, Maxwell Forbes, and Yejin Choi.
\newblock The curious case of neural text degeneration.
\newblock In {\em 8th International Conference on Learning Representations,
  {ICLR} 2020, Addis Ababa, Ethiopia, April 26-30, 2020}. OpenReview.net, 2020.

\bibitem{hu2022expansionnet}
Jia~Cheng Hu, Roberto Cavicchioli, and Alessandro Capotondi.
\newblock Expansionnet v2: Block static expansion in fast end to end training
  for image captioning.
\newblock {\em arXiv preprint arXiv:2208.06551}, 2022.

\bibitem{hu22lemon}
Xiaowei Hu, Zhe Gan, Jianfeng Wang, Zhengyuan Yang, Zicheng Liu, Yumao Lu, and
  Lijuan Wang.
\newblock Scaling up vision-language pretraining for image captioning.
\newblock In {\em CVPR}, 2022.

\bibitem{hudson19gqa}
Drew~A. Hudson and Christopher~D. Manning.
\newblock {GQA:} a new dataset for compositional question answering over
  real-world images.
\newblock In {\em CVPR}, 2019.

\bibitem{kim21vilt}
Wonjae Kim, Bokyung Son, and Ildoo Kim.
\newblock {ViLT}: Vision-and-language transformer without convolution or region
  supervision.
\newblock In {\em ICML}, 2021.

\bibitem{kirillov2019panoptic}
Alexander Kirillov, Kaiming He, Ross Girshick, Carsten Rother, and Piotr
  Doll{\'a}r.
\newblock Panoptic segmentation.
\newblock In {\em Proceedings of the IEEE/CVF Conference on Computer Vision and
  Pattern Recognition}, pages 9404--9413, 2019.

\bibitem{Kokkinos17UberNet}
Iasonas Kokkinos.
\newblock Ubernet: Training a universal convolutional neural network for low-,
  mid-, and high-level vision using diverse datasets and limited memory.
\newblock In {\em CVPR}, 2017.

\bibitem{kolesnikov2020bit}
Alexander Kolesnikov, Lucas Beyer, Xiaohua Zhai, Joan Puigcerver, Jessica Yung,
  Sylvain Gelly, and Neil Houlsby.
\newblock {Big Transfer ({BiT}): General Visual Representation Learning}.
\newblock In {\em ECCV}, 2020.

\bibitem{kolesnikov22uvim}
Alexander Kolesnikov, André~Susano Pinto, Lucas Beyer, Xiaohua Zhai,
  Jeremiah~J. Harmsen, and Neil Houlsby.
\newblock {UViM}: A unified modeling approach for vision with learned guiding
  codes.
\newblock In {\em NeurIPS}, 2022.

\bibitem{kool19gumbel}
Wouter Kool, Herke Van~Hoof, and Max Welling.
\newblock Stochastic beams and where to find them: The gumbel-top-k trick for
  sampling sequences without replacement.
\newblock In {\em International Conference on Machine Learning}, pages
  3499--3508. PMLR, 2019.

\bibitem{Krizhevsky12imagenet}
Alex Krizhevsky, Ilya Sutskever, and Geoffrey~E Hinton.
\newblock Imagenet classification with deep convolutional neural networks.
\newblock In {\em NeurIPS}, 2012.

\bibitem{Kuang17deepmulti}
Zhenzhong Kuang, Zongmin Li, Tianyi Zhao, and Jianping Fan.
\newblock Deep multi-task learning for large-scale image classification.
\newblock In {\em BigMM}, 2017.

\bibitem{kudo2018sentencepiece}
Taku Kudo and John Richardson.
\newblock {S}entence{P}iece: A simple and language independent subword
  tokenizer and detokenizer for neural text processing.
\newblock In {\em EMNLP}, 2018.

\bibitem{kurin22in}
Vitaly Kurin, Alessandro De~Palma, Ilya Kostrikov, Shimon Whiteson, and
  M.~Pawan Kumar.
\newblock In defense of the unitary scalarization for deep multi-task learning.
\newblock In {\em NeurIPS}, 2022.

\bibitem{lee19set}
Juho Lee, Yoonho Lee, Jungtaek Kim, Adam Kosiorek, Seungjin Choi, and Yee~Whye
  Teh.
\newblock Set transformer: A framework for attention-based
  permutation-invariant neural networks.
\newblock In Kamalika Chaudhuri and Ruslan Salakhutdinov, editors, {\em
  Proceedings of the 36th International Conference on Machine Learning},
  volume~97 of {\em Proceedings of Machine Learning Research}, pages
  3744--3753. PMLR, 09--15 Jun 2019.

\bibitem{li-etal-2022-mplug}
Chenliang Li, Haiyang Xu, Junfeng Tian, Wei Wang, Ming Yan, Bin Bi, Jiabo Ye,
  He Chen, Guohai Xu, Zheng Cao, Ji Zhang, Songfang Huang, Fei Huang, Jingren
  Zhou, and Luo Si.
\newblock m{PLUG}: Effective and efficient vision-language learning by
  cross-modal skip-connections.
\newblock In {\em Proceedings of the 2022 Conference on Empirical Methods in
  Natural Language Processing}, Dec. 2022.

\bibitem{li2022blip2}
Junnan Li, Dongxu Li, Silvio Savarese, and Steven Hoi.
\newblock Blip-2: Bootstrapping language-image pre-training with frozen image
  encoders and large language models.
\newblock {\em arXiv preprint arXiv:2301.12597}, 2023.

\bibitem{li2022blip}
Junnan Li, Dongxu Li, Caiming Xiong, and Steven Hoi.
\newblock {BLIP}: Bootstrapping language-image pre-training for unified
  vision-language understanding and generation.
\newblock In {\em ICML}, 2022.

\bibitem{li21albef}
Junnan Li, Ramprasaath~R. Selvaraju, Akhilesh~Deepak Gotmare, Shafiq Joty,
  Caiming Xiong, and Steven Hoi.
\newblock Align before fuse: Vision and language representation learning with
  momentum distillation.
\newblock In {\em NeurIPS}, 2021.

\bibitem{li19visualbert}
Liunian~Harold Li, Mark Yatskar, Da Yin, Cho-Jui Hsieh, , and Kai-Wei Chang.
\newblock {VisualBERT}: A simple and performant baseline for vision and
  language.
\newblock {\em arXiv preprint arXiv:1908.03557}, 2019.

\bibitem{li22explore}
Yanghao Li, Hanzi Mao, Ross Girshick, and Kaiming He.
\newblock Exploring plain vision transformer backbones for object detection.
\newblock {\em arXiv preprint arXiv:2203.16527}, 2022.

\bibitem{lin14mscoco}
Tsung-Yi Lin, Michael Maire, Serge Belongie, James Hays, Pietro Perona, Deva
  Ramanan, Piotr Doll{\'a}r, and C.~Lawrence Zitnick.
\newblock Microsoft coco: Common objects in context.
\newblock In {\em ECCV}, 2014.

\bibitem{liu21swin}
Ze Liu, Yutong Lin, Yue Cao, Han Hu, Yixuan Wei, Zheng Zhang, Stephen Lin, and
  Baining Guo.
\newblock Swin transformer: Hierarchical vision transformer using shifted
  windows.
\newblock 2021.

\bibitem{liu2022convnet}
Zhuang Liu, Hanzi Mao, Chao-Yuan Wu, Christoph Feichtenhofer, Trevor Darrell,
  and Saining Xie.
\newblock A {ConvNet} for the 2020s.
\newblock In {\em CVPR}, 2022.

\bibitem{Loshchilov19adamw}
Ilya Loshchilov and Frank Hutter.
\newblock Decoupled weight decay regularization.
\newblock In {\em ICLR}, 2019.

\bibitem{lu2019vilbert}
Jiasen Lu, Dhruv Batra, Devi Parikh, and Stefan Lee.
\newblock {ViLBERT}: Pretraining task-agnostic visiolinguistic representations
  for vision-and-language tasks.
\newblock In {\em NeurIPS}, 2019.

\bibitem{lu23unifiedio}
Jiasen Lu, Christopher Clark, Rowan Zellers, Roozbeh Mottaghi, and Aniruddha
  Kembhavi.
\newblock {UNIFIED-IO}: A unified model for vision, language, and multi-modal
  tasks.
\newblock In {\em ICLR}, 2023.

\bibitem{lu20twelve}
Jiasen Lu, Vedanuj Goswami, Marcus Rohrbach, Devi Parikh, and Stefan Lee.
\newblock 12-in-1: Multi-task vision and language representation learning.
\newblock In {\em Proceedings of the IEEE/CVF Conference on Computer Vision and
  Pattern Recognition (CVPR)}, June 2020.

\bibitem{Mishra19ocrvqa}
Anand Mishra, Shashank Shekhar, Ajeet~Kumar Singh, and Anirban Chakraborty.
\newblock {OCR-VQA: Visual Question Answering by Reading Text in Images}.
\newblock In {\em ICDAR}, 2019.

\bibitem{Silberman12depth}
Pushmeet~Kohli Nathan~Silberman, Derek~Hoiem and Rob Fergus.
\newblock Indoor segmentation and support inference from rgbd images.
\newblock In {\em ECCV}, 2012.

\bibitem{parkhi12a}
Omkar~M. Parkhi, Andrea Vedaldi, Andrew Zisserman, and C.~V. Jawahar.
\newblock Cats and dogs.
\newblock In {\em IEEE Conference on Computer Vision and Pattern Recognition},
  2012.

\bibitem{piergiovanni2022pre}
AJ Piergiovanni, Weicheng Kuo, and Anelia Angelova.
\newblock Pre-training image-language transformers for open-vocabulary tasks.
\newblock {\em arXiv preprint arXiv:2209.04372}, 2022.

\bibitem{requeima2019cnaps}
James Requeima, Jonathan Gordon, John Bronskill, Sebastian Nowozin, and
  Richard~E Turner.
\newblock Fast and flexible multi-task classification using conditional neural
  adaptive processes.
\newblock In {\em NeurIPS}. 2019.

\bibitem{Olga15ILSVRC}
Olga Russakovsky, Jia Deng, Hao Su, Jonathan Krause, Sanjeev Satheesh, Sean Ma,
  Zhiheng Huang, Andrej Karpathy, Aditya Khosla, Michael Bernstein,
  Alexander~C. Berg, and Li Fei-Fei.
\newblock {ImageNet Large Scale Visual Recognition Challenge}.
\newblock {\em IJCV}, 115(3):211--252, 2015.

\bibitem{ruder17survey}
Ruder Sebastian.
\newblock An overview of multi-task learning in deep neural networks.
\newblock {\em arXiv preprint arXiv:1706.05098}, 2017.

\bibitem{shao2019objects365}
Shuai Shao, Zeming Li, Tianyuan Zhang, Chao Peng, Gang Yu, Xiangyu Zhang, Jing
  Li, and Jian Sun.
\newblock Objects365: A large-scale, high-quality dataset for object detection.
\newblock In {\em Proceedings of the IEEE/CVF international conference on
  computer vision}, 2019.

\bibitem{shen2022association}
Jiayi Shen, Zehao Xiao, Xiantong Zhen, Cees G.~M. Snoek, and Marcel Worring.
\newblock Association graph learning for multi-task classification with
  category shifts.
\newblock In {\em NeurIPS}, 2022.

\bibitem{steiner22how}
Andreas~Peter Steiner, Alexander Kolesnikov, Xiaohua Zhai, Ross Wightman, Jakob
  Uszkoreit, and Lucas Beyer.
\newblock How to train your {ViT}? data, augmentation, and regularization in
  vision transformers.
\newblock {\em Transactions on Machine Learning Research}, 2022.

\bibitem{strudel2021segmenter}
Robin Strudel, Ricardo~Garcia Pinel, Ivan Laptev, and Cordelia Schmid.
\newblock Segmenter: Transformer for semantic segmentation.
\newblock In {\em ICCV}, 2021.

\bibitem{su2020vlbert}
Weijie Su, Xizhou Zhu, Yue Cao, Bin Li, Lewei Lu, Furu Wei, and Jifeng Dai.
\newblock {VL-BERT}: Pre-training of generic visual-linguistic representations.
\newblock In {\em ICLR}, 2019.

\bibitem{szegedy15inception}
C. Szegedy, W. Liu, Y. Jia, P. Sermanet, S. Reed, D. Anguelov, D. Erhan, V.
  Vanhoucke, and A. Rabinovich.
\newblock Going deeper with convolutions.
\newblock In {\em CVPR}, 2015.

\bibitem{tan21efficnet2}
Mingxing Tan and Quoc~V. Le.
\newblock {EfficientNetV2}: Smaller models and faster training.
\newblock In {\em ICML}, 2021.

\bibitem{xm3600}
Ashish~V. Thapliyal, Jordi Pont{-}Tuset, Xi Chen, and Radu Soricut.
\newblock Crossmodal-3600: {A} massively multilingual multimodal evaluation
  dataset.
\newblock In Yoav Goldberg, Zornitsa Kozareva, and Yue Zhang, editors, {\em
  Proceedings of the 2022 Conference on Empirical Methods in Natural Language
  Processing, {EMNLP} 2022, Abu Dhabi, United Arab Emirates, December 7-11,
  2022}, pages 715--729. Association for Computational Linguistics, 2022.

\bibitem{thapliyal22crossmodal}
Ashish~V. Thapliyal, Jordi Pont{-}Tuset, Xi Chen, and Radu Soricut.
\newblock Crossmodal-3600: {A} massively multilingual multimodal evaluation
  dataset.
\newblock In {\em EMNLP}, 2022.

\bibitem{touvron21training}
Hugo Touvron, Matthieu Cord, Matthijs Douze, Francisco Massa, Alexandre
  Sablayrolles, and Herve Jegou.
\newblock Training data-efficient image transformers \& distillation through
  attention.
\newblock In {\em ICML}, 2021.

\bibitem{Touvron2022DeiTIR}
Hugo Touvron, Matthieu Cord, and Herve Jegou.
\newblock {DeiT III: Revenge of the ViT}.
\newblock In {\em ECCV}, 2022.

\bibitem{van2017neural}
Aaron Van Den~Oord, Oriol Vinyals, et~al.
\newblock Neural discrete representation learning.
\newblock In {\em Advances in neural information processing systems}, 2017.

\bibitem{Vandenhende22multi}
S. Vandenhende, S. Georgoulis, W.~Van Gansbeke, M. Proesmans, D. Dai, and
  L.~Van Gool.
\newblock Multi-task learning for dense prediction tasks: A survey.
\newblock {\em IEEE Transactions on Pattern Analysis \& Machine Intelligence},
  44(07):3614--3633, 2022.

\bibitem{Vaswani17attention}
Ashish Vaswani, Noam Shazeer, Niki Parmar, Jakob Uszkoreit, Llion Jones,
  Aidan~N Gomez, \L~ukasz Kaiser, and Illia Polosukhin.
\newblock Attention is all you need.
\newblock In I. Guyon, U.~Von Luxburg, S. Bengio, H. Wallach, R. Fergus, S.
  Vishwanathan, and R. Garnett, editors, {\em NeurIPS}.

\bibitem{vedantam2015cider}
Ramakrishna Vedantam, C Lawrence~Zitnick, and Devi Parikh.
\newblock Cider: Consensus-based image description evaluation.
\newblock In {\em CVPR}, 2015.

\bibitem{vijayakumar16diverse}
Ashwin~K Vijayakumar, Michael Cogswell, Ramprasath~R Selvaraju, Qing Sun,
  Stefan Lee, David Crandall, and Dhruv Batra.
\newblock Diverse beam search: Decoding diverse solutions from neural sequence
  models.
\newblock {\em arXiv preprint arXiv:1610.02424}, 2016.

\bibitem{wang21ufo}
Jianfeng Wang, Xiaowei Hu, Zhe Gan, Zhengyuan Yang, Xiyang Dai, Zicheng Liu,
  Yumao Lu, and Lijuan Wang.
\newblock {UFO}: A unified transformer for vision-language representation
  learning.
\newblock {\em arXiv preprint arXiv:2111.10023}, 2021.

\bibitem{wang22git}
Jianfeng Wang, Zhengyuan Yang, Xiaowei Hu, Linjie Li, Kevin Lin, Zhe Gan,
  Zicheng Liu, Ce Liu, and Lijuan Wang.
\newblock {GIT}: A generative image-to-text transformer for vision and
  language.
\newblock {\em Transactions on Machine Learning Research}, 2022.

\bibitem{wang22ofa}
Peng Wang, An Yang, Rui Men, Junyang Lin, Shuai Bai, Zhikang Li, Jianxin Ma,
  Chang Zhou, Jingren Zhou, and Hongxia Yang.
\newblock {OFA}: Unifying architectures, tasks, and modalities through a simple
  sequence-to-sequence learning framework.
\newblock In {\em ICML}, 2022.

\bibitem{wang22beitv3}
Wenhui Wang, Hangbo Bao, Li Dong, Johan Bjorck, Zhiliang Peng, Qiang Liu, Kriti
  Aggarwal, Owais~Khan Mohammed, Saksham Singhal, Subhojit Som, and Furu Wei.
\newblock Image as a foreign language: {BEiT} pretraining for all vision and
  vision-language tasks.
\newblock {\em arXiv preprint arXiv:2208.10442}, 2022.

\bibitem{wang21pvt}
Wenhai Wang, Enze Xie, Xiang Li, Deng{-}Ping Fan, Kaitao Song, Ding Liang, Tong
  Lu, Ping Luo, and Ling Shao.
\newblock Pyramid vision transformer: {A} versatile backbone for dense
  prediction without convolutions.
\newblock In {\em ICCV}, 2021.

\bibitem{wang2021not}
Yulin Wang, Rui Huang, Shiji Song, Zeyi Huang, and Gao Huang.
\newblock Not all images are worth 16x16 words: Dynamic transformers for
  efficient image recognition.
\newblock In {\em NeurIPS}, 2021.

\bibitem{wang22simvlm}
Zirui Wang, Jiahui Yu, Adams~Wei Yu, Zihang Dai, Yulia Tsvetkov, and Yuan Cao.
\newblock {SimVLM}: Simple visual language model pretraining with weak
  supervision.
\newblock In {\em ICLR}, 2022.

\bibitem{weber2021deeplab2}
Mark Weber, Huiyu Wang, Siyuan Qiao, Jun Xie, Maxwell~D Collins, Yukun Zhu,
  Liangzhe Yuan, Dahun Kim, Qihang Yu, Daniel Cremers, et~al.
\newblock Deeplab2: A tensorflow library for deep labeling.
\newblock {\em arXiv preprint arXiv:2106.09748}, 2021.

\bibitem{wu16nmt}
Yonghui Wu, Mike Schuster, Zhifeng Chen, Quoc~V Le, Mohammad Norouzi, Wolfgang
  Macherey, Maxim Krikun, Yuan Cao, Qin Gao, Klaus Macherey, et~al.
\newblock Google's neural machine translation system: Bridging the gap between
  human and machine translation.
\newblock {\em arXiv preprint arXiv:1609.08144}, 2016.

\bibitem{xue-etal-2021-mt5}
Linting Xue, Noah Constant, Adam Roberts, Mihir Kale, Rami Al-Rfou, Aditya
  Siddhant, Aditya Barua, and Colin Raffel.
\newblock m{T}5: A massively multilingual pre-trained text-to-text transformer.
\newblock In {\em Proceedings of the 2021 Conference of the North American
  Chapter of the Association for Computational Linguistics: Human Language
  Technologies}, pages 483--498, Online, June 2021. Association for
  Computational Linguistics.

\bibitem{yang22temporally}
Shusheng Yang, Xinggang Wang, Yu Li, Yuxin Fang, Jiemin Fang, Wenyu Liu, Xun
  Zhao, and Ying Shan.
\newblock Temporally efficient vision transformer for video instance
  segmentation.
\newblock {\em arXiv preprint arXiv:2204.08412}, 2022.

\bibitem{ye22invpt}
Hanrong Ye and Dan Xu.
\newblock Invpt: Inverted pyramid multi-task transformer for dense scene
  understanding.
\newblock In {\em ECCV}, 2022.

\bibitem{young14flickr}
Peter Young, Alice Lai, Micah Hodosh, and Julia Hockenmaier.
\newblock From image descriptions to visual denotations: New similarity metrics
  for semantic inference over event descriptions.
\newblock {\em Transactions of the Association for Computational Linguistics},
  2:67--78, 2014.

\bibitem{yu2022coca}
Jiahui Yu, Zirui Wang, Vijay Vasudevan, Legg Yeung, Mojtaba Seyedhosseini, and
  Yonghui Wu.
\newblock Coca: Contrastive captioners are image-text foundation models.
\newblock {\em Transactions on Machine Learning Research}, 2022.

\bibitem{yu2022k}
Qihang Yu, Huiyu Wang, Siyuan Qiao, Maxwell Collins, Yukun Zhu, Hartwig Adam,
  Alan Yuille, and Liang-Chieh Chen.
\newblock {k-means Mask Transformer}.
\newblock In {\em European Conference on Computer Vision}, 2022.

\bibitem{yuan21florence}
Lu Yuan, Dongdong Chen, Yi-Ling Chen, Noel Codella, Xiyang Dai, Jianfeng Gao,
  Houdong Hu, Xuedong Huang, Boxin Li, Chunyuan Li, et~al.
\newblock Florence: A new foundation model for computer vision.
\newblock {\em arXiv preprint arXiv:2111.11432}, 2021.

\bibitem{Zamir2018Taskonomy}
Amir~R. Zamir, Alexander Sax, William Shen, Leonidas~J. Guibas, Jitendra Malik,
  and Silvio Savarese.
\newblock Taskonomy: Disentangling task transfer learning.
\newblock In {\em CVPR}, 2018.

\bibitem{zhai22scaling}
Xiaohua Zhai, Alexander Kolesnikov, Neil Houlsby, and Lucas Beyer.
\newblock Scaling vision transformers.
\newblock In {\em CVPR}, 2022.

\bibitem{siglip}
Xiaohua Zhai, Basil Mustafa, Alexander Kolesnikov, and Lucas Beyer.
\newblock Sigmoid loss for language image pre-training.
\newblock {\em arXiv preprint arXiv:2303.15343}, 2023.

\bibitem{zhang17survey}
Yu Zhang and Qiang Yang.
\newblock A survey on multi-task learning.
\newblock {\em arXiv preprint arXiv:1707.08114}, 2017.

\bibitem{zhou2017places}
Bolei Zhou, Agata Lapedriza, Aditya Khosla, Aude Oliva, and Antonio Torralba.
\newblock Places: A 10 million image database for scene recognition.
\newblock {\em IEEE Transactions on Pattern Analysis and Machine Intelligence},
  2017.

\bibitem{zhu22uni}
Xizhou Zhu, Jinguo Zhu, Hao Li, Xiaoshi Wu, Xiaogang Wang, Hongsheng Li,
  Xiaohua Wang, and Jifeng Dai.
\newblock {Uni-Perceiver}: Pre-training unified architecture for generic
  perception for zero-shot and few-shot tasks.
\newblock In {\em CVPR}, 2022.

\bibitem{zou2022xdecoder}
Xueyan Zou, Zi-Yi Dou, Jianwei Yang, Zhe Gan, Linjie Li, Chunyuan Li, Xiyang
  Dai, Jianfeng Wang, Lu Yuan, Nanyun Peng, Lijuan Wang, Yong~Jae Lee, and
  Jianfeng Gao.
\newblock Generalized decoding for pixel, image and language.
\newblock {\em arXiv preprint arXiv:2212.11270}, 2022.

\end{thebibliography}
